\documentclass{article}
\usepackage{graphicx}

\usepackage[main, final]{neurips_2026}

\usepackage[utf8]{inputenc} 
\usepackage[T1]{fontenc}    
\usepackage{hyperref}       
\usepackage{url}            
\usepackage{booktabs}       
\usepackage{amsfonts}       
\usepackage{nicefrac}       
\usepackage{microtype}      
\usepackage{xcolor}         
\usepackage{amsmath} 
\usepackage{multirow} 
\usepackage{amsmath} 
\usepackage{multirow} 
\usepackage{algorithm}
\usepackage{algpseudocode}
\usepackage{enumitem}
\usepackage{pgf}
\usepackage{tikz}
\usetikzlibrary{shapes.geometric, arrows.meta, positioning, calc}
\usepackage{xurl}

\title{LEAD: Length-Efficient Adaptive and Dynamic Reasoning for Large Language Models}

\author{%
  Songtao Wei$^{1}$, Yi Li$^{1}$, Zhikai Li$^{2}$, Xu Hu$^{1}$, Yuede Ji$^{4}$, Guanpeng Li$^{5}$,
  Feng Chen$^{1}$\\ 
  \textbf{Carl Yang$^{2}$,
  Zhichun Guo$^{3}$,
  Bingzhe Li$^{1}$}\\
   \\
  $^1$University of Texas at Dallas \quad $^2$Emory University  \\
  $^3$Individual Researcher \quad $^4$University of Texas at Arlington  \\
  $^5$University of Florida
}

\begin{document}

\maketitle

\begin{abstract}
Large reasoning models, such as OpenAI o1 and DeepSeek-R1, tend to become increasingly verbose as their reasoning capabilities improve. These inflated Chain-of-Thought (CoT) trajectories often exceed what the underlying problems require, wasting compute, latency, and context budgets. While introducing length-based efficiency rewards during reinforcement learning offers a natural remedy, existing methods struggle with two fundamental challenges: the optimal balance between correctness and efficiency is non-stationary throughout training, and intrinsic reasoning budgets vary drastically across problems. Relying on static reward weights and global length constraints inevitably forces a compromise between degraded accuracy and unrealized compression. To overcome these limitations, we propose LEAD (\textbf{L}ength-\textbf{E}fficient \textbf{A}daptive and \textbf{D}ynamic reasoning), a method that replaces static heuristics with online, self-adaptive mechanisms. LEAD dynamically calibrates the correctness--efficiency trade-off at each step using a Potential-Scaled Instability, directing optimization capacity to the most informative learning signal. Furthermore, it estimates an adaptive per-problem target length online based on the model’s own correct rollouts, applying a symmetric efficiency reward that penalizes both overthinking and over-compression. 
Evaluated on five mathematical reasoning benchmarks, LEAD achieves the highest accuracy and Accuracy-Efficiency Score among RL-trained efficient-reasoning methods in our matched comparisons while producing substantially shorter outputs than the base model.

\end{abstract}

\section{Introduction}
\label{introduction}

Chain-of-thought (CoT) prompting~\cite{wei2022chain} shows that large language models (LLMs) can improve complex problem solving through explicit intermediate reasoning, inspiring many subsequent reasoning and tool-use methods~\cite{wang2022self,zhou2022least,yao2023tree,besta2024graph,yao2022react,schick2023toolformer}.
More recently, reinforcement learning (RL) has further strengthened reasoning models such as OpenAI o1~\cite{jaech2024openai} and DeepSeek-R1~\cite{guo2025deepseek}, producing long and elaborate reasoning traces that improve performance on challenging tasks.
However, this emergent reasoning comes at a cost: reasoning models are verbose by default. As models improve, their solutions grow longer, consuming compute, latency, and context budget on reasoning steps that are often unnecessary for the problem at hand~\cite{chen2024not, chen2025tokenflow}. A competition-level math problem may legitimately require thousands of reasoning tokens; a single-step arithmetic query should not. Yet models trained solely to maximize correctness learn to ``think longer to think better,'' producing responses whose length is largely decoupled from the complexity of the underlying task.

Making LLM reasoning \emph{efficient} has therefore become a central research question~\cite{
arora2025training, xiang2025just, aggarwal2025l1, luo2025o1,
yi2025shorterbetter, he2025smartthinker, li2025drpo, liu2025learn,
li2025selfbudgeter, shrivastava2025sample}. The standard recipe is to augment the RL training loop with a length-based efficiency signal in addition to the correctness signal, either through reward shaping~\cite{arora2025training, yi2025shorterbetter, he2025smartthinker, team2025kimi, liu2025learn}, multi-objective reinforcement learning~\cite{li2025drpo, huang2025hapo, aggarwal2025l1, liu2026gdpo, shrivastava2025sample, lu2025learning}, or trajectory-level constraints~\cite{hou2025thinkprune, yu2025dapo, luo2025o1, li2025selfbudgeter, muennighoff2025s1}.
In principle, this signal should encourage the model to remove redundant reasoning while preserving the reasoning needed for correctness. 
In practice, however, this goal depends on two questions that static length-control schemes do not answer well: \emph{when} should the optimizer prioritize brevity during training, and \emph{how much} reasoning should each problem be allowed to use? 

These questions expose two challenges that efficient reasoning methods must address. 
\textbf{The first challenge is to dynamically balance reward contributions over training.} 
The relative usefulness of rewards for correctness and efficiency changes as the policy improves.
Early in training, correctness-oriented exploration is essential, and excessive length pressure can suppress reasoning needed to discover valid solutions. 
As training progresses and some prompts become reliably solvable, the efficiency signal becomes more useful for removing redundant reasoning from those solved trajectories. 
Thus, a fixed reward ratio $(\lambda_c,\lambda_\ell)$ is unlikely to remain appropriate throughout training. 
\textbf{The second challenge is adaptive efficiency across problem difficulties.} 
Different prompts require different amounts of reasoning, so a single target length should not be applied uniformly across all problems.
A simple arithmetic problem may be solved concisely, whereas an Olympiad-level problem may require many intermediate steps.
A global budget either over-compresses hard problems, hurting correctness, or under-compresses easy problems, wasting tokens.
Together, these challenges call for a framework that \emph{dynamically} balances reward contributions throughout training while \emph{adaptively} calibrating the target length for each prompt.

We propose \textbf{LEAD} (\textbf{L}ength-\textbf{E}fficient \textbf{A}daptive and \textbf{D}ynamic reasoning), a framework that addresses both challenges through online self-calibration. 
LEAD combines two mechanisms. 
\emph{First}, it dynamically adjusts the correctness--efficiency trade-off during training. Rewards are normalized separately to prevent scale dominance, and their weights are updated online according to which signal remains informative. 
This creates a transient curriculum in which length efficiency guides early compression, while optimization gradually shifts toward correctness as the efficiency signal saturates. 
\emph{Second}, LEAD replaces a global length budget with a per-prompt target $L^*_q$ estimated from the model's current correct rollouts. 
This target adapts to both problem difficulty and model capability, allowing hard prompts to retain the necessary reasoning while encouraging easy prompts to be concise. 
A symmetric efficiency reward around $L^*_q$ penalizes both overthinking and over-compression.

We evaluate LEAD on five math reasoning benchmarks using different LLM models. LEAD matches or exceeds baseline accuracy while significantly reducing solution length, outperforming recent efficient-reasoning methods (DRPO~\cite{li2025drpo}, ShorterBetter~\cite{yi2025shorterbetter}) on the accuracy--efficiency score. Our contributions are:
\begin{itemize}
    \item We identify two algorithm-agnostic challenges in efficient-reasoning RL: dynamic reward balancing over training and adaptive efficiency across problem difficulties, and show they are difficult to resolve reliably with a static coefficient without task- and model-specific tuning.
    \item We propose LEAD, which combines online instability-driven reward weighting with per-problem target-length calibration, requiring no manual coefficient scheduling.
    \item We validate LEAD across five math benchmarks and on 1.5B- and 7B-sized models, showing consistent improvements in the accuracy--efficiency trade-off over state-of-the-art baselines. The code is released~\footnote{\url{https://github.com/CrazyMint/LEAD}.}.
\end{itemize}

\section{Related Work}
\label{related_work}

\paragraph{Reinforcement Learning for LLM Reasoning.}
Outcome-based reinforcement learning is the dominant paradigm for training large reasoning models such as OpenAI o1~\cite{jaech2024openai}, DeepSeek-R1~\cite{guo2025deepseek}, Kimi-k1.5~\cite{team2025kimi}, and Qwen-QwQ~\cite{qwen2.5}, all of which scale test-time chain-of-thought to deliver substantial gains on complex reasoning tasks. The most widely used algorithm in this setting is GRPO~\cite{guo2025deepseek}, which samples multiple rollouts per prompt and computes group-relative advantages under a clipped policy-gradient objective without a critic. 
DAPO~\cite{yu2025dapo} extends GRPO with dynamic sampling, token-level policy gradients, and overlong reward shaping for large-scale stability. 
More generally, optimizing reasoning for both correctness and efficiency is a multi-objective RL problem, where simple scalarization can obscure trade-offs between competing objectives~\cite{hayes2022practical}.
When multiple reward signals are combined, GDPO~\cite{liu2026gdpo} identifies reward-advantage collapse in GRPO's combine-then-normalize design, where the higher-variance signal dominates after normalization, and mitigates it by normalizing each reward separately before combining them with static weights.

\paragraph{Efficient Reasoning.}
A growing body of work addresses the verbosity problem in reasoning models, namely the tendency to generate unnecessarily long solutions when optimized primarily for correctness.
Several methods introduce length penalties, pruning objectives, or budget constraints during training.
L1~\cite{aggarwal2025l1} trains reasoning models to follow user-specified length constraints, O1-Pruner~\cite{luo2025o1} uses length-harmonizing fine-tuning to reduce redundant long-thought reasoning, and DRPO~\cite{li2025drpo} decouples the learning signals for correct and incorrect rollouts to avoid penalizing valid long reasoning.
LASER~\cite{liu2025learn} formulates efficient reasoning through adaptive length-based reward shaping, while GFPO~\cite{shrivastava2025sample} encourages concise reasoning by filtering sampled rollouts according to length and reward-per-token efficiency.
LASER uses a target-length step reward, and LASER-D additionally adapts targets to training progress and difficulty. LEAD combines an online per-prompt mean-correct target, a symmetric deviation reward, and reward-conditioned weighting of separately normalized advantages.
Other methods estimate or impose problem-dependent budgets:
ShorterBetter~\cite{yi2025shorterbetter} uses the shortest correct rollout as a Sample Optimal Length,
SmartThinker~\cite{he2025smartthinker} calibrates reasoning length through a distributional estimate,
SelfBudgeter~\cite{li2025selfbudgeter} predicts query-specific token budgets before generation,
and e1~\cite{kleinman2025e1} learns adaptive control of reasoning effort through an inference-time effort parameter.
A complementary line studies test-time compute allocation rather than training-time reward optimization: s1~\cite{muennighoff2025s1} uses budget forcing for test-time scaling, Plan-and-Budget~\cite{lin2025plan} allocates token budgets across decomposed subproblems, and Agarwal et al.~\cite{agarwal2025art} show that the best test-time scaling strategy depends on model type, problem difficulty, and compute budget.

\paragraph{Trajectory editing and learned reasoning control.}
LC-R1~\cite{cheng2026lcr1} combines a length reward with a compression reward aimed at redundant thinking after an answer is found. JET~\cite{han2026jet} uses truncated training rollouts and a quality-controlled length reward. Step-GRPO~\cite{chen2026stepgrpo} combines dynamic rollout truncation with semantic-step-relative rewards to internalize early exit. These methods modify the exploration trajectories or the unit of length control, whereas LEAD calibrates a token-length target from ordinary capped rollouts. Think-How-to-Think~\cite{liu2026th2t} uses staged training for difficulty and redundancy awareness, and Correct, Concise and Complete~\cite{rakotonirina2026c3} combines supervised preparation with RL that penalizes continuation beyond the first correct answer. TIME~\cite{das2026time} instead learns context-triggered reasoning in temporally structured dialogue. These approaches offer complementary ways to learn when reasoning is needed. LEAD focuses on online reward allocation and behavior-relative target estimation within verifiable mathematical reasoning.


\section{Limitations of Static Length Control}
\label{sec:limitation}

\paragraph{Notation.} We consider a reasoning policy $\pi_\theta$ trained on a dataset $\mathcal{D}=\{q_i\}_{i=1}^{N}$ of prompts. Following GRPO~\cite{guo2025deepseek}, for each prompt $q$ we sample a group of $G$ rollouts $\{o_{q,j}\}_{j=1}^{G}$ from the old policy $\pi_{\theta_\text{old}}$, each with token length $\ell_{q,j} = |o_{q,j}|$. Let $r_c(o,q) \in \{0,1\}$ denote the binary correctness reward and $r_\ell(o, q) \in \mathbb{R}$ a length-based efficiency reward. In standard GRPO, the final reward is a scalar combination $r(o,q) = \lambda_c \, r_c(o,q) + \lambda_\ell \, r_\ell(o,q)$ with non-negative weights $\lambda_c, \lambda_\ell$ (their relative ratio $\rho = \lambda_\ell/\lambda_c$ controls how much the optimizer listens to length), and the group-relative advantage $A_{q,j}$ is shared across all tokens $t$ of rollout $j$:

\begin{equation}
\label{eq:grpo_adv}
A_{q,j} \;=\; \frac{r(o_{q,j}, q) - \mu_q}{\sigma_q + \epsilon},\qquad
\mu_q = \tfrac{1}{G}\sum_{j} r(o_{q,j},q),\ \ \sigma_q = \mathrm{Std}_j\, r(o_{q,j},q).
\end{equation}
The policy is then updated by minimizing a loss that is the negative of the clipped PPO-style surrogate over $A_{q,j}$ plus a KL regularizer (full objective deferred to Appendix~\ref{app:grpo_loss}). While this formulation works well for a single reward, the combined-then-normalized structure of Eq.~\eqref{eq:grpo_adv} introduces structural pathologies when applied to jointly optimize accuracy and efficiency. We identify two such pathologies below, both of which motivate our method.

\subsection{Reward Collapse under Static Weighting}
\label{sec:reward_collapse}
The group normalization in Eq.~\eqref{eq:grpo_adv} is applied \emph{after} the two reward components have already been combined. Consider a group in which all $G$ rollouts are correct ($r_c{=}1$) and differ only in length. The combined reward reduces to $r(o_{q,j},q) = \lambda_c + \lambda_\ell\, r_\ell(o_{q,j},q)$, so $\mu_q = \lambda_c + \lambda_\ell\, \mu_q^{(\ell)}$ and $\sigma_q = \lambda_\ell\, \sigma_q^{(\ell)}$. 
Substituting into Eq.~\eqref{eq:grpo_adv}, for any $\lambda_\ell > 0$ and ignoring the numerical regularizer $\epsilon$, the static trade-off coefficient cancels in numerator and denominator and the advantage reduces to $A_{q,j} \approx (r_\ell(o_{q,j},q) - \mu_q^{(\ell)})/\sigma_q^{(\ell)}$: the length penalty drives the gradient at full normalized magnitude regardless of the practitioner's intended $\lambda_\ell$.
Conversely, in an all-incorrect group ($r_c{=}0$), the same cancellation means the efficiency signal drives the entire advantage, even though there is no correctness to preserve. 
In mixed groups, the scale mismatch between binary correctness and continuous length rewards causes the higher-variance component to dominate after normalization, while the other becomes noise. 
Tuning static weights cannot fully solve this, because the useful balance changes over training. Length rewards are informative while the model is learning to compress, but their within-group variance collapses once responses cluster, whereas correctness often remains informative on hard prompts. 
Thus, a fixed pair $(\lambda_c,\lambda_\ell)$ either over-compresses before solving is learned or underuses length feedback after accuracy stabilizes.


\subsection{Global Length Budget Ignores Problem Difficulty}
\label{sec:global_budget}

A second limitation is how the efficiency reward itself is shaped. 
A common strategy applies a global length budget $B$ to all prompts~\cite{yu2025dapo, li2025drpo, hou2025thinkprune, team2025kimi}, e.g., $r_\ell=\min(0,1-\ell/B)$ once the response exceeds the budget. 

This ignores the heterogeneity of reasoning difficulty. For example, easy arithmetic and olympiad-level problems should not share the same target length.
When $B$ is set aggressively to drive compression, the model is forced to truncate its reasoning on hard problems that genuinely require more steps, producing short but often incorrect outputs. This is a well-documented accuracy regression in prior efficient-reasoning methods~\cite{arora2025training, huang2025hapo, li2025drpo}. When $B$ is set loosely to preserve accuracy, the penalty rarely fires on easy problems, and the compression benefit vanishes. Thus, a fixed global budget cannot simultaneously respect problem-dependent reasoning requirements and exploit compression opportunities when they exist. 
Both failure modes arise from the same mismatch: a single global budget cannot reflect the heterogeneous reasoning demands of different prompts.

\section{Method}
\label{method}

\begin{figure}[t]
    \centering
    \includegraphics[width=\linewidth]{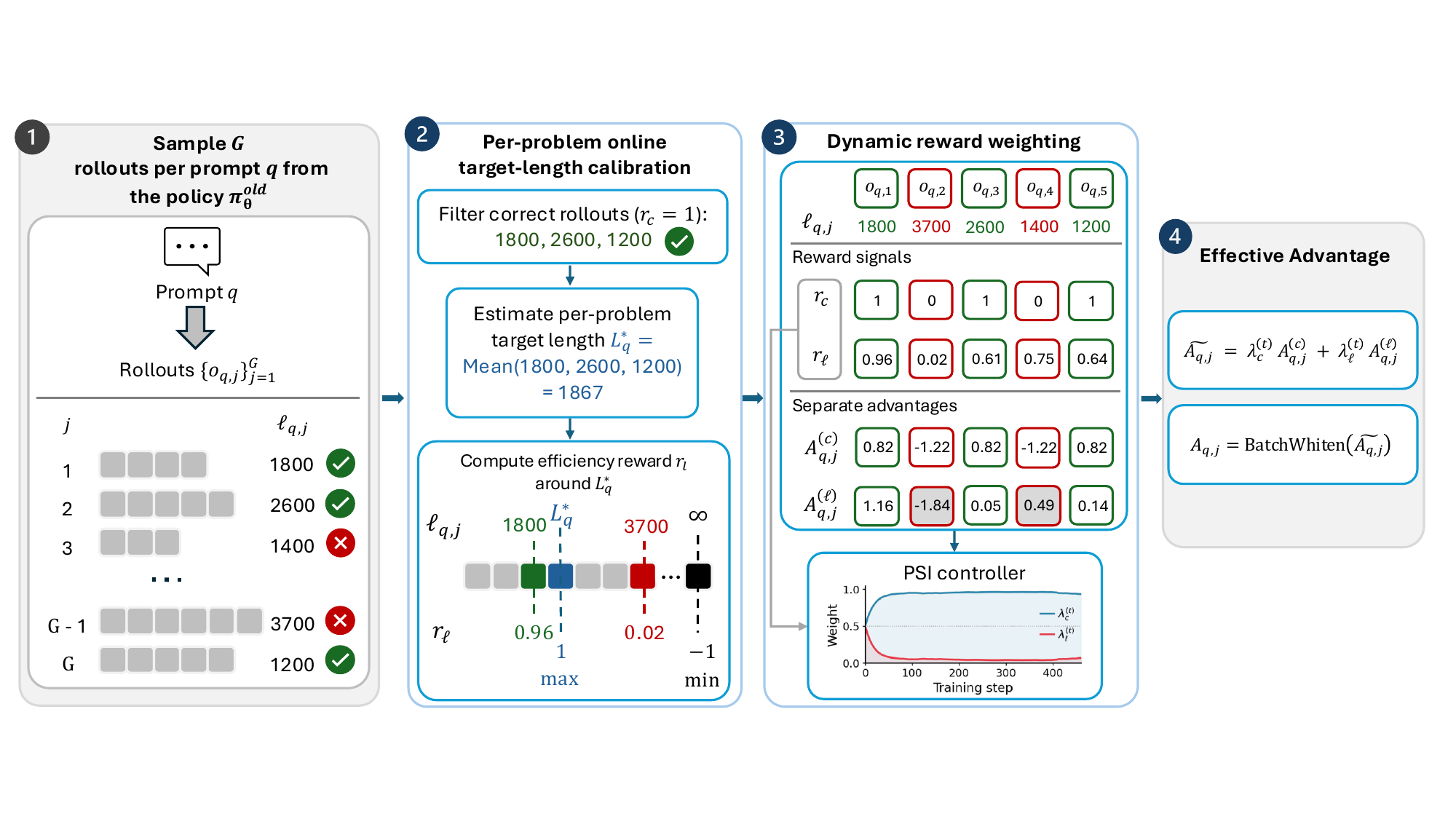}
    \vspace{-25pt}
    \caption{\textbf{LEAD framework.} 
    \textbf{(1)}~Sample $G$ rollouts per prompt $q$ from the old policy $\pi_{\theta_\text{old}}$ and score each by correctness $r_c$. 
    \textbf{(2)}~Per-problem online target-length calibration: filter to correct rollouts $\mathcal{C}_q$, set $L^*_q$ to their mean, and compute the symmetric efficiency reward $r_\ell$, which peaks at $\ell{=}L^*_q$ and decays linearly to $-1$ on either side. 
    \textbf{(3)}~Dynamic reward weighting: each reward is group-normalized separately to produce decoupled advantages $A^{(c)}_{q,j}$ and $A^{(\ell)}_{q,j}$, and the PSI controller adapts the weights $(\lambda_c^{(t)}, \lambda_\ell^{(t)})$ online from per-reward instability and headroom. 
    \textbf{(4)}~The two advantage channels are linearly combined under the EMA-smoothed weights and batch-whitened to obtain the effective advantage $A_{q,j}$ used in the GRPO objective. 
    \textcolor[RGB]{0,155,230}{Blue-framed} components are LEAD's contributions; \textcolor{gray}{gray-framed} components are inherited from GRPO.}
    \label{fig:lead}
\end{figure}

LEAD has two key components: \textbf{dynamic reward weighting with decoupled group normalization} (Section~\ref{sec:dynamic_weighting}), which combines per-reward normalized advantages under online, instability-driven weights instead of the scalar-combined advantage of Eq.~\eqref{eq:grpo_adv}; and \textbf{per-problem online target-length calibration} (Section~\ref{sec:online_budget}), which replaces the global length budget with a per-problem target $L^*_q$ estimated from the model's own correct rollouts. Figure~\ref{fig:lead} shows the full pipeline.

\subsection{Dynamic Reward Weighting with Decoupled Group Normalization}
\label{sec:dynamic_weighting}

\paragraph{Decoupled group normalization.} Following GDPO~\cite{liu2026gdpo}, we normalize each reward in its own group before aggregation, which prevents the reward-advantage collapse of Section~\ref{sec:reward_collapse}. For each reward $k \in \{c, \ell\}$,
\begin{equation}
\label{eq:decoupled_adv}
A^{(k)}_{q,j} \;=\; \frac{r_k(o_{q,j},q) - \mu_q^{(k)}}{\sigma_q^{(k)} + \epsilon},\qquad
\mu_q^{(k)} = \tfrac{1}{G}\sum_j r_k(o_{q,j},q),\ \ \sigma_q^{(k)} = \mathrm{Std}_j\, r_k(o_{q,j},q),
\end{equation}
and the components are combined under a weight vector $\boldsymbol{\lambda} = (\lambda_c, \lambda_\ell)$ with $\lambda_k \geq 0$, $\sum_k \lambda_k = 1$:
\begin{equation}
\label{eq:combined_adv}
\tilde{A}_{q,j} \;=\; \lambda_c\, A^{(c)}_{q,j} \;+\; \lambda_\ell\, A^{(\ell)}_{q,j},\qquad
A_{q,j} \;=\; \mathrm{BatchWhiten}(\tilde{A}_{q,j}) \;=\; \frac{\tilde{A}_{q,j} - \bar{\mu}}{\sqrt{\bar{v} + \epsilon}},
\end{equation}
where $\bar{\mu}$ and $\bar v$ are the mean and sample variance of $\tilde{A}$ over valid response-token positions after broadcasting each rollout advantage to its tokens (Appendix~\ref{app:whiten_detail}). Because each $A^{(k)}$ is already group-centered, $\bar\mu$ is close to zero and BatchWhiten mainly rescales the combined advantage to unit variance across the batch. We keep the explicit centering for numerical robustness. Decoupled normalization addresses only the scale-mismatch half of the pathology in Section~\ref{sec:reward_collapse}: it prevents the reward with the larger within-group variance from drowning out the other, but it inherits GDPO's assumption that a fixed $(\lambda_c, \lambda_\ell)$ is appropriate throughout training. The non-stationary half remains, since the relative learnability of the two rewards drifts as one saturates faster than the other. We close this gap with online dynamic weighting.

\paragraph{Dynamic weighting via the Potential-Scaled Instability (PSI).} With scale mismatch already removed by decoupled normalization, the remaining question is which reward still provides a usable learning signal at the current training step. We adapt $\boldsymbol{\lambda}$ online from two statistics of each reward: its \emph{instability} (a reward still changing rapidly carries a gradient signal) and its \emph{headroom} (a reward near its ceiling cannot improve further). At each training step, from the current batch of $M$ prompts ($G$ rollouts each), the Law of Total Variance gives the raw-reward mean and standard deviation as
\begin{equation}
\label{eq:total_var}
\mu_k \;=\; \tfrac{1}{M}\sum_{q=1}^{M} \mu_q^{(k)},\qquad
\sigma_k \;=\; \sqrt{\tfrac{1}{M}\sum_{q=1}^{M} \mathrm{Var}_j\, r_k(o_{q,j},q) \;+\; \mathrm{Var}_q\, \mu_q^{(k)} \;+\; \epsilon},
\end{equation}
where $\mathrm{Var}_j$ and $\mathrm{Var}_q$ are population variances over the rollouts of a prompt and over the prompt means of the batch, and the coefficient of variation $\mathrm{CV}_k = \sigma_k / (|\mu_k| + \epsilon)$ measures instability relative to magnitude.\footnote{For the efficiency reward ($k{=}\ell$), the per-prompt mean and variance entering Eq.~\eqref{eq:total_var} are restricted to correct rollouts $\mathcal{C}_q$, and prompts with $|\mathcal{C}_q|{=}0$ are dropped from the outer average, since incorrect rollouts carry no usable efficiency signal. The per-rollout advantage in Eq.~\eqref{eq:decoupled_adv} continues to use all $G$ rollouts. The $\epsilon{=}10^{-8}$ regularizer in the CV denominator handles transient zero-crossings during early warmup.} The \textbf{potential} $P_k$ measures headroom to the reward's ceiling, given the reward's range $[R_k^{\min}, R_k^{\max}]$ ($[0,1]$ for correctness; $[-1,1]$ for our symmetric length reward):
\begin{equation}
\label{eq:potential}
P_k \;=\; \left(1 - \frac{\mu_k - R_k^{\min}}{R_k^{\max} - R_k^{\min}}\right)^{\alpha},
\end{equation}
where $\alpha$ controls the decay sharpness near the ceiling. The combined \textbf{potential-scaled instability (PSI)} is
\begin{equation}
\label{eq:psi}
\Psi_k \;=\; \widetilde{\mathrm{CV}}_k \cdot P_k,\qquad
\widetilde{\mathrm{CV}}_k \;=\; \frac{\mathrm{CV}_k}{\sum_{k'} \mathrm{CV}_{k'} + \epsilon},
\end{equation}
which is large when the reward $k$ is noisy and far from the ceiling; small when stable or saturated.


\paragraph{Why $\widetilde{\mathrm{CV}}_k \cdot P_k$.}
After decoupled normalization removes scale mismatch, a reward should receive high weight only if it remains both informative and improvable.
$\widetilde{\mathrm{CV}}_k$ measures relative reward variability, while $P_k$ measures remaining headroom to the reward ceiling.
The two factors capture orthogonal failure modes: a reward can have ample variance yet sit near its ceiling on most prompts, or have substantial headroom but little usable variation across rollouts.
Their product is large only when both conditions hold and small when either fails.
Unlike GradNorm~\cite{chen2018gradnorm} or uncertainty weighting~\cite{kendall2018multi}, which balance raw gradient or loss scales, PSI uses inexpensive batch statistics of the raw rewards: group normalization assigns relative credit within each prompt, while PSI sets the ratio between the two reward channels from their batch-level variability and headroom.

Per-batch $\Psi_k$ values are noisy, so we normalize and EMA-smooth them into the target weights:
\begin{equation}
\label{eq:ema}
\hat\lambda_k^{(t)} = \frac{\Psi_k}{\sum_{k'} \Psi_{k'} + \epsilon},\qquad
\boldsymbol{u}^{(t)} \;=\; \beta_\mathrm{ema}\, \boldsymbol{u}^{(t-1)} \;+\; (1-\beta_\mathrm{ema})\, \hat{\boldsymbol{\lambda}}^{(t)},\qquad \boldsymbol{u}^{(0)} = \mathbf{1}/K,
\end{equation}
with $\beta_\mathrm{ema}=0.95$ (effective horizon of roughly 20 steps). The smoothed state $\boldsymbol u^{(t)}$ is stored before any floor is applied, and the applied weights enforce a floor $\lambda_c \geq \lambda_{\min}$ by setting $\lambda_c^{(t)}=\max(u_c^{(t)},\lambda_{\min})$ and $\lambda_\ell^{(t)}=1-\lambda_c^{(t)}$, which preserves $\sum_k \lambda_k = 1$. This prevents the correctness signal from being fully dampened by a transiently stable batch. Fixed reward bounds keep the potential well defined when every rollout has the same reward, and Appendix~\ref{app:psi_edges} describes the all-correct and all-wrong cases. The only added state beyond GRPO is $\boldsymbol u^{(t)}$ (two scalars). The full procedure is summarized in Algorithm~\ref{alg:lead}.

\subsection{Per-problem Online Target-Length Calibration}
\label{sec:online_budget}

The second component replaces the global budget $B$ with a per-problem target length $L^*_q$ estimated per prompt from the model's own correct rollouts, addressing the heterogeneity and over-compression issues of Section~\ref{sec:global_budget}.

\paragraph{Online target-length estimation.} Let $\mathcal{C}_q = \{j : r_c(o_{q,j},q) = 1\}$ be the indices of correct rollouts for prompt $q$. We define $L^*_q$ as the mean length of $\mathcal{C}_q$, clamped to a permissible range:
\begin{equation}
\label{eq:lstar}
L^*_q \;=\;
\begin{cases}
\mathrm{clip}\!\left(\tfrac{1}{|\mathcal{C}_q|}\sum_{j \in \mathcal{C}_q} \ell_{q,j},\ L_{\min},\ B_{\max}\right) & \text{if } |\mathcal{C}_q| \geq 1, \\[4pt]
B_{\max} & \text{if } |\mathcal{C}_q| = 0,
\end{cases}
\end{equation}
where $L_{\min}$ keeps the reward well-conditioned for very short solutions and $B_{\max}$ is the training-time max response length, doubling as the upper clamp and the sentinel value for unsolved prompts. When $|\mathcal{C}_q|=0$, setting $L^*_q = B_{\max}$ makes Eq.~\eqref{eq:reff_sym} reduce to $r_\ell = \ell_{q,j}/B_{\max}$, which after group normalization places the longest rollouts in the group at positive efficiency advantage and the shortest at negative. We accept this expansion pressure on unsolved prompts as a deliberate trade-off: correctness on those prompts is what matters first, so encouraging longer reasoning while the model is still searching for a solution is consistent with the long-reasoning behavior already present in the base model. Its contribution to the policy gradient is small in practice because $\lambda_\ell$ is small in steady state (Appendix~\ref{app:dynamics} reports $\lambda_\ell \approx 0.07$ post-warmup) and the fraction of unsolved prompts diminishes as training progresses. $L^*_q$ adapts to both prompt and model: harder prompts produce longer correct rollouts and larger $L^*_q$, and as the model learns to solve a problem more concisely, $L^*_q$ tightens automatically, sustaining compression without a manual curriculum. Using the \emph{mean} rather than the \emph{minimum} of correct lengths (contrast with ShorterBetter's SOL~\cite{yi2025shorterbetter}) prevents a single anomalously short rollout from setting an unrealistically aggressive target. Appendix~\ref{app:sparse_targets} examines sparse-correct groups and the zero-correct fallback.

\paragraph{Symmetric efficiency reward.} Given $L^*_q$, the efficiency reward is symmetric around the target:
\begin{equation}
\label{eq:reff_sym}
r_\ell(o_{q,j},q) \;=\; \max\!\left(-1,\ 1 - \frac{\bigl|\ell_{q,j} - L^*_q\bigr|}{L^*_q}\right).
\end{equation}
The reward equals $1$ at $\ell_{q,j} = L^*_q$, decreases linearly with deviation, and is clipped at $-1$. Penalizing under-length is intentional: an over-short ``correct'' solution often signals a shortcut (pattern-matched answer) rather than reasoning, and rewarding it would reintroduce the over-compression pathology of Section~\ref{sec:global_budget}. Because $L^*_q$ is recomputed each batch from the current correct rollouts, the penalty on a genuinely short-but-valid solution is transient rather than permanent: if the policy actually discovers shorter solutions on prompt $q$, those rollouts pull $L^*_q$ downward in subsequent updates, and the symmetric form tracks the new optimum. Appendix~\ref{app:target_dynamics} examines this target evolution and its interaction with the training cap.

\paragraph{Rollout masks for advantages and controller statistics.} The per-rollout efficiency advantage $A^{(\ell)}_{q,j}$ in Eq.~\eqref{eq:decoupled_adv} and the batch-level controller statistics in Eq.~\eqref{eq:total_var} use different masks, which we list explicitly. \emph{(i) Per-rollout advantages.} For every prompt $q$, $r_\ell$ is computed for all $G$ rollouts from the prompt's $L^*_q$, and $\mu_q^{(\ell)}, \sigma_q^{(\ell)}$ are taken over the full group of $G$. An incorrect rollout whose length is near $L^*_q$ therefore receives a non-trivial efficiency advantage, which the correctness channel offsets through its negative correctness advantage. Using all $G$ rollouts also avoids the singular case $|\mathcal{C}_q|{=}1$, where a within-correct-only standard deviation would be zero. \emph{(ii) Controller statistics.} For the PSI controller only, the per-prompt $\mu_q^{(\ell)}, \sigma_q^{(\ell)}$ are computed over $\mathcal{C}_q$ (a single correct rollout contributes its reward as $\mu_q^{(\ell)}$ with $\sigma_q^{(\ell)}{=}0$), and the outer average runs over the $M' \leq M$ prompts with $|\mathcal{C}_q|{\geq}1$. Prompts with $|\mathcal{C}_q|{=}0$ are dropped because their efficiency reward $r_\ell = \ell_{q,j}/B_{\max}$ carries no usable signal about target-length tracking. This mask affects only the global controller weight $\lambda_\ell$, not the per-rollout advantage. Without it, unsolved prompts would inflate $\mathrm{CV}_\ell$ early in training and spuriously up-weight efficiency before correctness is achieved (edge cases are listed in Appendix~\ref{app:psi_edges}).

\section{Experiment}
\label{experiment}

\subsection{Experimental Setup}
\label{sec:experimental_setup}

\paragraph{Dataset and Models.}
We train all methods on the level 3--5 split of the MATH dataset~\cite{hendrycks2021measuring}, comprising $8{,}521$ problems. We use DeepSeek-R1-Distill-Qwen-1.5B and DeepSeek-R1-Distill-Qwen-7B~\cite{guo2025deepseek} as base models. The MATH-500 benchmark used at evaluation time is held out and disjoint from the training pool.

\paragraph{Baselines.}
We evaluate LEAD against state-of-the-art RL baselines for LLM reasoning and correctness--length optimization.
\textbf{GRPO~\cite{guo2025deepseek}} aggregates rewards into a single scalar before group-wise normalization, serving as our foundation for standard and statically-weighted multi-objective training.
\textbf{GDPO~\cite{liu2026gdpo}} mitigates scale dominance by normalizing objectives independently before aggregation, but relies on a static combination of reward weights.
\textbf{DRPO~\cite{li2025drpo}} decouples learning signals via a discriminative framework, applying a fixed, global length penalty exclusively to correct responses to encourage conciseness without inverting validity.
\textbf{ShorterBetter~\cite{yi2025shorterbetter}} penalizes deviation from a dynamic Sample Optimal Length (SOL) target, but uses fixed-weight GRPO, lacking the ability to dynamically shift optimization focus as accuracy stabilizes.

\paragraph{Implementation.}
We implement LEAD on top of the Verl framework~\cite{sheng2024hybridflow} with vLLM~\cite{kwon2023efficient} as the rollout engine. Training hyperparameters (optimizer, batch sizes, KL, PPO clip, $G$, epochs) and LEAD's controller settings are listed in Appendix~\ref{app:hyperparams}, Table~\ref{tab:hyperparams}. The 7B runs use smaller batches and a lower learning rate than the 1.5B runs but share the rollout, clip, and LEAD blocks. We set $L_{\min}{=}1{,}000$ because the training distribution is MATH Level 3--5, where even the easiest correct rollouts span several hundred to a few thousand tokens. Other method-specific hyperparameters follow the original baselines, and Appendix~\ref{app:protocol_detail} records the baseline optimizer settings. 

\paragraph{Evaluation.}
We evaluate all models with five math reasoning benchmarks: AIME 2024, AIME 2025, AMC 2023, MATH-500~\cite{hendrycks2021measuring}, 
and OlympiadBench~\cite{he2024olympiadbench}. Following the sampling setup of Sober Reasoning~\cite{hochlehnert2025sober}, we use temperature $0.8$, top-$p=0.9$, and $n=10$ decodes per question for AIME24/25 and AMC23, or $n=3$ for MATH-500 and OlympiadBench. \textbf{Accuracy} is avg@$n$: the mean correctness of the $n$ decodes, not the probability that at least one succeeds (pass@$n$). \textbf{Average Length} includes all responses, correct or incorrect, and excludes prompt tokens. Both headline metrics weight the five benchmark means equally. Evaluation uses a 32,768-token context setting, distinct from the 4,000/8,000-token training caps. Generation and truncation details appear in Appendix~\ref{app:protocol_detail}. \textbf{AES}~\cite{luo2025o1} combines relative accuracy and length changes against the same-scale base, using the coefficients of DRPO~\cite{li2025drpo} (Appendix~\ref{app:aes}).

\subsection{Results}
\label{sec:main_results}


\paragraph{Math Reasoning Results.}
Table~\ref{tab:main_results} reports per-benchmark accuracy, response length, and AES for models trained with a max response length of $4,000$. 
Across both 1.5B and 7B scales, LEAD achieves the highest average accuracy and AES among all trained methods in Table~\ref{tab:main_results}. 
For the 1.5B model, LEAD is the only method that improves over the base model while reducing average length, reaching $53.36$ accuracy and $0.68$ AES. Compared with the strongest baseline DRPO, it improves accuracy by $2.62$ points and AES by $0.18$. 
For the 7B model, all RL-trained methods reduce length but regress from the base model in accuracy. LEAD has the smallest accuracy drop and the best AES, reaching $65.17$ accuracy and $-0.11$ AES. 
Although LEAD is not the shortest method, its stronger AES indicates a better accuracy--efficiency trade-off: it preserves more reasoning tokens when they are useful, rather than uniformly compressing all prompts. 
This is consistent with the per-problem target design in Section~\ref{sec:online_budget} and the difficulty-based allocation analysis in Appendix~\ref{app:difficulty}.

\begin{table}[t]
\centering
\caption{Performance comparison across methods on DeepSeek-R1-Distill-Qwen-1.5B and -7B trained with a 4K maximum response length. AES denotes the Accuracy-Efficiency Score. For each model, \textbf{bold} marks the best Acc / AES among trained methods and \underline{underline} marks the second best.}
\label{tab:main_results}
\resizebox{\textwidth}{!}{%
\begin{tabular}{l cc cc cc cc cc | cc c}
\toprule
\multirow{2}{*}{\textbf{Model}} & \multicolumn{2}{c}{\textbf{AIME24}} & \multicolumn{2}{c}{\textbf{AIME25}} & \multicolumn{2}{c}{\textbf{AMC23}} & \multicolumn{2}{c}{\textbf{MATH}} & \multicolumn{2}{c|}{\textbf{OlyBch}} & \multicolumn{2}{c}{\textbf{Average}} & \textbf{AES~$\uparrow$} \\
\cmidrule(lr){2-3} \cmidrule(lr){4-5} \cmidrule(lr){6-7} \cmidrule(lr){8-9} \cmidrule(lr){10-11} \cmidrule(lr){12-13} \cmidrule(lr){14-14}
 & Acc & Len & Acc & Len & Acc & Len & Acc & Len & Acc & Len & Acc & Len &  \\
\midrule
\multicolumn{14}{c}{\textit{DeepSeek-R1-Distill-Qwen-1.5B}} \\
\midrule
Base            & 29.33 & 13018 & 24.00 & 13126 & 69.25 & 7472 & 84.93 & 4080 & 51.75 & 8370 & 51.85 & 9213 & -- \\
\midrule
GRPO            & 23.67 & 2723 & 19.67 & 2227 & 65.75 & 1634 & 82.27 &  903 & 49.23 & 1747 & 48.12 & 1847 &  0.08 \\
GDPO            & 20.00 & 2232 & 16.67 & 1837 & 61.25 & 1545 & 80.80 &  957 & 48.74 & 1645 & 45.49 & 1643 & $-$0.40 \\
ShorterBetter   & 24.67 & 5361 & 18.67 & 5649 & 65.00 & 2685 & 80.40 & 1466 & 44.84 & 3159 & 46.72 & 3664 & $-$0.39 \\
DRPO            & \underline{27.33} & 3658 & \underline{21.00} & 3745 & \textbf{70.25} & 2173 & \underline{83.27} & 1254 & \underline{51.85} & 2404 & \underline{50.74} & 2647 & \underline{0.50} \\
\textbf{LEAD (Ours)} & \textbf{35.00} & 5133 & \textbf{24.33} & 4550 & \underline{67.50} & 3336 & \textbf{85.47} & 2100 & \textbf{54.52} & 3450 & \textbf{53.36} & 3714 & \textbf{0.68} \\
\midrule
\multicolumn{14}{c}{\textit{DeepSeek-R1-Distill-Qwen-7B}} \\
\midrule
Base            & 53.67 & 9444 & 39.00 & 10263 & 87.75 & 4552 & 93.53 & 2717 & 67.06 & 6181 & 68.20 & 6631 & -- \\
\midrule
GRPO            & 37.67 & 4245 & 27.67 & 3421 & 83.25 & 1714 & 88.93 &  979 & 59.01 & 2049 & 59.31 & 2482 & $-$0.68 \\
GDPO            & 41.33 & 3159 & 27.67 & 2871 & 80.25 & 1551 & 89.73 &  940 & 60.05 & 2335 & 59.81 & 2171 & $-$0.56 \\
ShorterBetter   & \textbf{46.33} & 4765 & \underline{31.33} & 5345 & 80.75 & 1583 & 83.53 &  741 & 60.49 & 2489 & 60.49 & 2985 & $-$0.58 \\
DRPO            & 44.00 & 5090 & 30.67 & 5232 & \underline{88.00} & 2255 & \underline{92.07} & 1334 & \textbf{65.23} & 3420 & \underline{63.99} & 3466 & \underline{$-$0.14} \\
\textbf{LEAD (Ours)} & \underline{44.67} & 6631 & \textbf{35.00} & 7090 & \textbf{88.75} & 2662 & \textbf{92.27} & 1705 & \underline{65.14} & 3997 & \textbf{65.17} & 4417 & \textbf{$-$0.11} \\
\bottomrule
\end{tabular}%
}
\end{table}

\subsection{Ablation Study}
\label{sec:ablation}

We isolate two central design choices of LEAD with controlled ablations on DeepSeek-R1-Distill-Qwen-1.5B trained on MATH (Level~3--5) under the same evaluation protocol as Section~\ref{experiment}: (i) static vs.\ dynamic reward weighting, paired with the orthogonal scalarized vs.\ decoupled normalization choice, and (ii) the choice of aggregator for the per-problem target length $L^*_q$. 

\paragraph{Static vs.\ Dynamic reward weighting.}
We compare scalarized GRPO, static decoupled weighting, and LEAD's dynamic weighting by sweeping six fixed ratios with $\lambda_c+\lambda_\ell=1$ and reporting dynamic LEAD as an independent reference. 
Table~\ref{tab:abl_dynamic} shows three trends. 
First, scalarized GRPO is highly sensitive to the reward ratio: its best AES is only $0.08$ at $1{:}1$, with most ratios at or below zero, indicating that combine-then-normalize aggregation makes the length signal difficult to control. 
Second, every static LEAD ratio achieves AES $\geq 0.27$, outperforming the best GRPO setting. Besides separately normalizing the two rewards, static LEAD applies the final token-masked BatchWhiten, which remains active even at $1{:}0$, and the GRPO sweep uses the baseline's optimizer settings, so the gap at $1{:}0$ reflects these pipeline differences rather than reward decoupling alone (Appendices~\ref{app:whiten_detail} and~\ref{app:protocol_detail}). 
Third, dynamic LEAD achieves the best overall trade-off without manually selecting a ratio, reaching AES $0.68$ and accuracy $53.36$, above the best static setting. 
The dynamic weights explain this behavior: $\lambda_\ell$ starts near $0.5$ and decays to $\approx 0.07$ as the efficiency signal saturates, allowing LEAD to first learn compression and then shift optimization toward correctness (shown in Appendix~\ref{app:dynamics}). Section~\ref{sec:additional_results} compares PSI with hand-designed schedules under a matched configuration.

\begin{table}[t]
\centering
\caption{Static vs.\ dynamic weighting on DeepSeek-R1-Distill-Qwen-1.5B (4K). Both \textbf{GRPO} and \textbf{LEAD} (static) use static combination $(\lambda_c,\lambda_\ell)$, and \textbf{LEAD (dynamic)} adopts $\lambda^{(t)}$ online (i.e., dynamic combination of $(\lambda_c,\lambda_\ell)$).}

\label{tab:abl_dynamic}
\small
\setlength{\tabcolsep}{4pt}
\begin{tabular}{c|ccc|ccc|ccc}
\toprule
\multirow{2}{*}{$\lambda_c{:}\lambda_\ell$}
& \multicolumn{3}{c|}{\textbf{GRPO}}
& \multicolumn{3}{c|}{\textbf{LEAD (static)}}
& \multicolumn{3}{c}{\textbf{LEAD (dynamic)}} \\
\cmidrule(lr){2-4} \cmidrule(lr){5-7} \cmidrule(lr){8-10}
& Acc & Len & $\mathrm{AES}\uparrow$
& Acc & Len & $\mathrm{AES}\uparrow$
& Acc & Len & $\mathrm{AES}\uparrow$ \\
\midrule
1 : 0  & 44.59 & 2320 & $-$0.65 & 50.60 & 3141 &  0.42 & \multirow{6}{*}{\textbf{53.36}} & \multirow{6}{*}{3714} & \multirow{6}{*}{\textbf{0.68}} \\
4 : 1  & 47.87 & 1802 &  0.04   & 52.74 & 4013 &  0.62 & & & \\
2 : 1  & 47.24 & 1689 & $-$0.07 & 52.16 & 3586 &  0.63 & & & \\
1 : 1  & 48.12 & 1847 &    0.08 & 51.46 & 5177 &  0.36 & & & \\
1 : 2  & 46.11 & 1527 & $-$0.27 & 52.26 & 5819 &  0.39 & & & \\
1 : 4  & 47.27 & 1670 & $-$0.07 & 51.08 & 5322 &  0.27 & & & \\
\bottomrule
\end{tabular}
\end{table}

\begin{figure}[t]
    \centering
    \includegraphics[width=\linewidth]{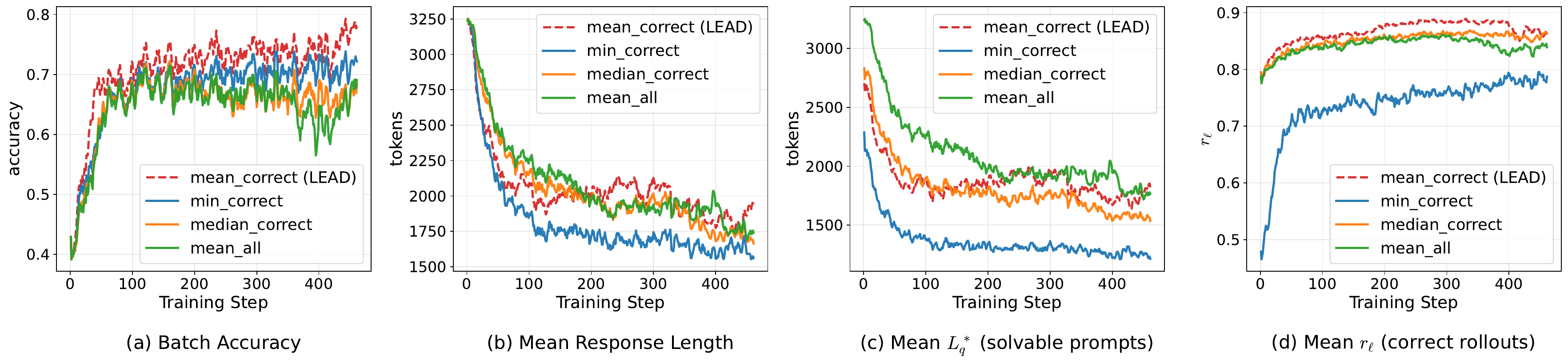}
    \vspace{-16pt}
    \caption{\textbf{Training trajectories of the four aggregator variants on DeepSeek-R1-Distill-Qwen-1.5B.} \textbf{(a)} On-policy batch accuracy. \textbf{(b)} Mean response length on the rollout batch. \textbf{(c)} Per-problem target $L^*_q$ averaged over solvable prompts. \textbf{(d)} Symmetric efficiency reward $r_\ell$ averaged over correct rollouts.}
    \label{fig:abl_aggregator_dyn}
\end{figure}

\paragraph{Choice of $L^*_q$ aggregator.}
We vary the aggregator: the mean of correct rollouts (LEAD), the minimum (ShorterBetter's SOL~\cite{yi2025shorterbetter}), the median, and a degenerate baseline averaging over \emph{all} rollouts regardless of correctness. As shown in Table~\ref{tab:abl_aggregator}, LEAD wins on both Acc (53.36) and AES (0.68). Min (Min of correct) compresses most aggressively, but a single short outlier sets an unrealistically tight target, causing accuracy to drop by over 3 points. Median (Median of correct) yields the worst AES (0.34): its length sits between Min and Mean-of-all, yet its accuracy is no better than Min's. Mean-of-all is the closest competitor, but trails LEAD by 1.9 accuracy points because the inflation from incorrect rollouts is absorbed by off-track or truncated trajectories rather than valid reasoning.
\begin{table}[t]
\centering
\caption{Choice of aggregator for $L^*_q$. All variants restrict to correct rollouts $\mathcal{C}_q$ except \emph{Mean of all rollouts}, the unfiltered baseline.}
\label{tab:abl_aggregator}
\small
\begin{tabular}{llccc}
\toprule
Aggregator & Definition & Avg Acc & Avg Len & $\mathrm{AES}\uparrow$ \\
\midrule
Min of correct~\cite{yi2025shorterbetter} & $\min_{j \in \mathcal{C}_q} \ell_{q,j}$ & 49.99 & 2611 & 0.36 \\
Median of correct & $\mathrm{med}_{j \in \mathcal{C}_q} \ell_{q,j}$ & 50.13 & 2990 & 0.34 \\
Mean of all rollouts & $\mathrm{mean}_j \ell_{q,j}$ & 51.47 & 3230 & 0.58 \\
\textbf{Mean of correct (LEAD)} & $\mathrm{mean}_{j \in \mathcal{C}_q} \ell_{q,j}$ & \textbf{53.36} & 3714 & \textbf{0.68} \\
\bottomrule
\end{tabular}
\end{table}
Figure~\ref{fig:abl_aggregator_dyn} explains this behavior through the training dynamics. LEAD pulls ahead in batch accuracy in (a) while Min lags. Rollout lengths in (b) converge for Median, LEAD, and Mean-of-all, while Min compresses further. The targets $L^*_q$ in (c) stay ordered Min$<$Median$<$LEAD$<$Mean-of-all. Mean-of-all's target runs $\sim$150--200 tokens above LEAD's, even though their rollout lengths converge in (b), so the gap is absorbed by incorrect rollouts rather than reshaping the target on solvable problems. The diagnostic is (d): Median, LEAD, and Mean-of-all all reach $r_\ell \approx 0.85$, so correct trajectories land near $L^*_q$. Min plateaus at $\approx 0.78$, since even its own correct rollouts cannot meet such a tight target, so the efficiency penalty fights correctness, explaining Min's lag in (a).

\subsection{Additional Results}
\label{sec:additional_results}
Full tables and protocol details for the following experiments are in Appendix~\ref{app:additional_evidence}.

\paragraph{Training-run variability.}
Re-training GRPO, GDPO, DRPO, and LEAD on DeepSeek-R1-Distill-Qwen-1.5B (4K) with three data-order seeds gives LEAD a mean accuracy of $52.93\pm0.91$ and AES of $0.67\pm0.05$, against $50.82\pm0.67$ and $0.50\pm0.12$ for DRPO, the strongest baseline (Table~\ref{tab:seed_summary}). LEAD's worst seed exceeds DRPO's three-seed mean. Paired question-bootstrap intervals for the Table~\ref{tab:main_results} models exclude zero at 1.5B but include zero at 7B (Appendix~\ref{app:uncertainty}).

\paragraph{Comparison with pre-defined weighting schedules.}
We compare our PSI controller with some pre-defined weighting schedule settings. A linear decay of $\lambda_\ell$ from $0.5$ to $0.05$ over the first 50 updates is competitive with PSI at step 300 and worse at step 450. A two-phase schedule that switches from equal weights to correctness only after 50 updates is also included in the comparison and is shown as the worst setting. PSI therefore exceeds a well-chosen schedule without specifying it in advance and continues to improve both accuracy and length later in training.

\paragraph{Comparison with released LASER-family checkpoints.}
Under the same evaluation protocol, LEAD is more accurate and shorter than all nine released 1.5B LASER, LASER-D, and LASER-DE checkpoints (best LASER AES $0.53$ versus $0.68$). At 7B, the two released adaptive checkpoints are more accurate and shorter than LEAD. Since the training protocols differ (LASER uses the larger DeepScaleR training dateset and a larger 16K cap), so these are comparisons of released models rather than matched training runs (Appendix~\ref{app:laser_comparison}). However, LEAD is still competitive with more restricted training setups.

\paragraph{Cross-family transfer with LoRA.}
We compare LEAD-LoRA and GRPO-LoRA on Qwen3-4B~\cite{yang2025qwen3}. GRPO combines correctness and its scalar length reward at a $1{:}1$ ratio. Both LEAD-LoRA and GRPO-LoRA share the same 4,000-token training cap and training horizon. In Table~\ref{tab:qwen3_lora_summary}, LEAD retains more accuracy than GRPO ($+3.06$ points) and achieves higher AES, while producing longer responses than GRPO and shortening the base model's responses by $30.5\%$. More details are shown in Appendix~\ref{app:qwen3_lora}.

\section{Conclusion}
\label{sec:conclusion}

We introduced \textbf{LEAD}, an online self-calibrating framework for length-efficient reasoning motivated by two challenges in RL-based reasoning optimization: the changing correctness--efficiency trade-off during training and the heterogeneous reasoning demands of different prompts. 
LEAD combines a Potential-Scaled Instability controller, which adapts $(\lambda_c,\lambda_\ell)$ from per-reward instability and remaining headroom, with a per-problem target length $L^*_q$ estimated from the model's own correct rollouts under a symmetric efficiency reward. 
Across five mathematical reasoning benchmarks, LEAD achieves the best Accuracy-Efficiency Score among GRPO, GDPO, DRPO, ShorterBetter, and their length-control variants, without requiring a hand-tuned $(\lambda_c,\lambda_\ell)$ schedule. 
Ablations show that dynamic weighting improves over the evaluated static reward ratios, and the mean-of-correct target outperforms the minimum, median, and correctness-blind alternatives. 
Overall, LEAD is a step toward reasoning models that adapt their computational footprint online to both problem difficulty and optimization progress, improving efficiency without sacrificing the reasoning performance.

\section{Acknowledgment}
This work was partially supported by NSF 2343863, 2413520, 2417747 and 2440611. Any opinions, conclusions, or recommendations expressed in this material are those of the authors and do not necessarily reflect the views of the NSF.

\newpage

\bibliography{references}
\bibliographystyle{unsrt}


\appendix

\section{Limitations and Broader Impact}
\label{app:limitations_impact}

\subsection{Limitations}
LEAD is designed for reinforcement learning settings where correctness can be reliably evaluated and multiple rollouts can be sampled per prompt. This makes it well suited to mathematical reasoning, where answer verification is relatively precise, but extending the same framework to open-ended generation, instruction following, or subjective preference tasks may require task-specific reward models or validators. We view this as a natural extension rather than a limitation of the core mechanism, since the dynamic weighting and per-problem calibration components are agnostic to the particular reward source.

LEAD estimates its target length from the model's own correct rollouts. This design intentionally prioritizes correctness before compression: prompts that the model cannot yet solve do not receive target-based compression pressure until at least one correct trajectory appears. As a result, efficiency gains on very hard prompts may emerge later in training than on easier prompts. However, this behavior is desirable for reasoning tasks, since premature compression on unsolved problems can suppress the exploration needed to discover valid solutions.

A related limitation is that the target is self-referential. Because it is estimated from the policy's own correct rollouts, it can stabilize above an unknown minimum sufficient reasoning length, particularly when successful trajectories cluster tightly around a behavioral floor. Within-group length variance may then shrink, PSI may down-weight the efficiency channel, and further compression can stall. Appendix~\ref{app:behavioral_floor} measures how often such tight high-length clusters occur in our runs. They are uncommon in the evaluated setting, but the possibility should be kept in mind when LEAD is applied to other models or domains. Targets estimated from only one or two correct rollouts can also be noisy (Appendix~\ref{app:sparse_targets}), and PSI uses reward variability and headroom as heuristic proxies for learnability rather than exact measures of policy-gradient utility. Our cross-family evidence is a Qwen3-4B LoRA study in which both trained methods lose accuracy relative to the base under the 4K cap, and transfer to larger models such as Qwen3-32B remains future work.

The current formulation uses a single scalar target length for each prompt, which is appropriate for math reasoning where correct solutions often cluster around a problem-dependent reasoning budget. Some tasks may admit multiple valid solution styles with substantially different lengths, such as concise direct answers and longer explanatory derivations. Extending LEAD from a single target to a distributional or multi-target length model is an interesting direction for future work.

Finally, LEAD improves length efficiency through training-time policy optimization rather than enforcing a hard inference-time token budget. Therefore, it should be viewed as a method for learning more efficient reasoning behavior, not as a replacement for deployment-time budget controllers when strict latency or cost constraints are required. Combining LEAD with test-time budget allocation may further improve efficiency under fixed resource limits.

\subsection{Broader Impact}
LEAD aims to reduce unnecessary reasoning tokens while preserving task performance. 
By shortening redundant reasoning trajectories, it can lower inference cost, latency, and energy use in deployments where reasoning models are queried at scale. 
Because LEAD is a reward-aggregation and length-calibration method, it does not directly introduce new task capabilities beyond those induced by the underlying RL training objective.
At the same time, improving reasoning efficiency may change the form of model outputs: shorter responses can be harder for users to inspect when detailed explanations are needed. 
For applications where transparency, debugging, or human verification is important, we recommend retaining longer reasoning traces, auxiliary logs, or configurable verbosity settings in controlled settings. 
More broadly, LEAD should be deployed with the same safeguards as the base reasoning model, since efficiency improvements can also make both beneficial and harmful uses cheaper to run.

\section{Full GRPO Objective}
\label{app:grpo_loss}

Following the notation of Section~\ref{sec:limitation}, the policy is updated by \emph{minimizing} the loss $\mathcal{L}_{\mathrm{GRPO}}$ defined below, which equals the negative of the clipped PPO-style surrogate over the group-relative advantage $A_{q,j}$ of Eq.~\eqref{eq:grpo_adv} plus a KL-divergence regularizer against a reference policy $\pi_{\text{ref}}$. Minimizing $\mathcal{L}_{\mathrm{GRPO}}$ therefore maximizes the surrogate while penalizing deviation from $\pi_{\text{ref}}$:
\begin{equation}
\label{eq:grpo_loss}
\mathcal{L}_{\mathrm{GRPO}}(\theta) \;=\; -\,\mathbb{E}_{q \sim \mathcal{D},\, \{o_{q,j}\} \sim \pi_{\theta_\text{old}}}\!\left[\frac{1}{G}\sum_{j=1}^{G}\frac{1}{|o_{q,j}|}\sum_{t=1}^{|o_{q,j}|}\!\mathcal{J}_{q,j,t}\right] + \beta\,\mathbb{D}_{\mathrm{KL}}\!\bigl[\pi_\theta \,\|\, \pi_{\text{ref}}\bigr],
\end{equation}
\begin{equation}
\label{eq:grpo_clip}
\begin{gathered}
\mathcal{J}_{q,j,t} \;=\; \min\!\Bigl(\rho_{q,j,t}(\theta)\, A_{q,j},\ \ \mathrm{clip}\bigl(\rho_{q,j,t}(\theta),\,1-\varepsilon,\,1+\varepsilon\bigr)\, A_{q,j}\Bigr), \\[2pt]
\rho_{q,j,t}(\theta) \;=\; \frac{\pi_\theta(o_{q,j,t}\mid q, o_{q,j,<t})}{\pi_{\theta_\text{old}}(o_{q,j,t}\mid q, o_{q,j,<t})},
\end{gathered}
\end{equation}
where $\rho_{q,j,t}(\theta)$ is the per-token importance ratio, $\varepsilon$ is the clipping threshold, and $\beta$ controls the KL penalty strength. LEAD inherits this objective unchanged; the only modification is the construction of $A_{q,j}$ described in Section~\ref{method}.

\section{LEAD Algorithm}
\label{app:algorithm}

Algorithm~\ref{alg:lead} summarizes the LEAD advantage computation at a single training step, combining the per-reward decoupled normalization (Section~\ref{sec:dynamic_weighting}), the PSI-driven dynamic weighting (Eqs.~\eqref{eq:psi}--\eqref{eq:ema}), and the per-problem online target length (Section~\ref{sec:online_budget}).

\begin{algorithm}[t]
\caption{LEAD advantage computation at training step $t$}
\label{alg:lead}
\begin{algorithmic}[1]
\Require Rollouts $\{o_{q,j}\}$, correctness $\{r_c(o_{q,j}, q)\}$, lengths $\{\ell_{q,j}\}$, prior pre-floor EMA state $\boldsymbol u^{(t-1)}$

\vspace{3pt}
\State \textbf{\textit{Phase 1: Online Target-Length Calibration \& Decoupled Normalization}}
\For{each prompt $q$ in the current batch}
    \State Compute per-prompt target $L^*_q$ from correct rollouts via Eq.~\eqref{eq:lstar}
    \State Compute symmetric efficiency rewards $r_\ell(o_{q,j},q)$ for all $j \in \{1\dots G\}$ via Eq.~\eqref{eq:reff_sym}
    \State Compute decoupled group-relative advantages $A^{(c)}_{q,j}$ and $A^{(\ell)}_{q,j}$ via Eq.~\eqref{eq:decoupled_adv}
\EndFor

\vspace{3pt}
\State \textbf{\textit{Phase 2: Dynamic PSI Controller}}
\State Compute batch-level reward statistics $\mu_k, \sigma_k$ via Eq.~\eqref{eq:total_var} (for $k{=}\ell$, restrict to correct rollouts $\mathcal{C}_q$ and drop prompts with $|\mathcal{C}_q|{=}0$; the per-rollout advantage in line~5 is unchanged and uses all $G$ rollouts)
\State Compute instability $\mathrm{CV}_k$, potential headroom $P_k$, and PSI $\Psi_k$ via Eqs.~\eqref{eq:potential}--\eqref{eq:psi}
\State Derive target weights $\hat{\boldsymbol{\lambda}}^{(t)}$, update and retain $\boldsymbol u^{(t)}$ via Eq.~\eqref{eq:ema}
\State Apply $\lambda_c^{(t)}=\max(u_c^{(t)},\lambda_{\min})$, $\lambda_\ell^{(t)}=1-\lambda_c^{(t)}$

\vspace{3pt}
\State \textbf{\textit{Phase 3: Advantage Aggregation}}
\State Combine signals into scalar advantage $\tilde{A}_{q,j} = \lambda_c^{(t)} A^{(c)}_{q,j} + \lambda_\ell^{(t)} A^{(\ell)}_{q,j}$
\State \Return Final advantages $A_{q,j} = \mathrm{BatchWhiten}(\tilde{A}_{q,j})$
\end{algorithmic}
\end{algorithm}

\section{Accuracy-Efficiency Score}
\label{app:aes}

The Accuracy-Efficiency Score~\cite{luo2025o1} jointly measures accuracy preservation and length reduction relative to a reference model (the base model before RL training):
\begin{equation}
\text{AES} =
\begin{cases}
\alpha \cdot \Delta_{\text{Len}} + \beta \cdot |\Delta_{\text{Acc}}|, & \text{if } \Delta_{\text{Acc}} \geq 0, \\
\alpha \cdot \Delta_{\text{Len}} - \gamma \cdot |\Delta_{\text{Acc}}|, & \text{if } \Delta_{\text{Acc}} < 0,
\end{cases}
\end{equation}
where $\Delta_{\text{Len}} = \frac{L_{\text{ref}} - L_{\text{model}}}{L_{\text{ref}}}$ measures relative length reduction, $\Delta_{\text{Acc}} = \frac{A_{\text{model}} - A_{\text{ref}}}{A_{\text{ref}}}$ measures relative accuracy change, and we use $\alpha{=}1$, $\beta{=}3$, $\gamma{=}10$ following~\cite{li2025drpo}, with a large $\gamma$ to emphasize the importance of minimizing accuracy degradation. Higher AES indicates a better accuracy--efficiency trade-off.

\section{Additional Experiment}
\label{app:additional-experiment}
This appendix collects full hyperparameter specifications and additional empirical results, including training dynamics and complementary results.

\subsection{Full Hyperparameter Specification}
\label{app:hyperparams}

Table~\ref{tab:hyperparams} records the reported LEAD configurations. It is not a universal configuration for all baseline rows: baseline optimizer, filtering, and horizon choices differ. Appendix~\ref{app:protocol_detail} distinguishes original comparisons from the matched schedule runs.

\begin{table}[h]
\centering
\caption{Reported LEAD training configurations for DeepSeek-R1-Distill-Qwen-1.5B and -7B. ``Shared'' refers to these two columns, not all baseline methods.}
\label{tab:hyperparams}
\small
\begin{tabular}{lcc}
\toprule
\textbf{Setting} & \textbf{1.5B} & \textbf{7B} \\
\midrule
\multicolumn{3}{l}{\emph{Optimizer / rollout}} \\
\midrule
Learning rate & $2\!\times\!10^{-6}$ & $1\!\times\!10^{-6}$ \\
Training batch size (prompts) & $128$ & $32$ \\
PPO mini-batch size & $16$ & $8$ \\
PPO micro-batch size & $16$ & $4$ \\
KL loss coefficient & $5\!\times\!10^{-4}$ & $1\!\times\!10^{-3}$ \\
Total epochs / steps & $7$ / $462$ & $2$ / $\sim$$532$ \\
Max response length $B_{\max}$ & $4{,}000$ (4K) / $8{,}000$ (8K) & $4{,}000$ \\
Rollouts per prompt $G$ & \multicolumn{2}{c}{$8$ (shared)} \\
Rollout temperature & \multicolumn{2}{c}{$1.0$ (shared)} \\
PPO epochs per update & \multicolumn{2}{c}{$1$ (shared)} \\
PPO clip ratio $\varepsilon$ & \multicolumn{2}{c}{$0.2$ (Verl default; ratio range $[0.8, 1.2]$)} \\
Max prompt length & \multicolumn{2}{c}{$1{,}024$ (shared)} \\
Outer KL controller coefficient & \multicolumn{2}{c}{$1\!\times\!10^{-3}$ (shared, low-variance KL)} \\
\midrule
\multicolumn{3}{l}{\emph{LEAD-specific (shared across scales)}} \\
\midrule
Potential exponent $\alpha$ & \multicolumn{2}{c}{$1.0$} \\
EMA smoothing $\beta_\mathrm{ema}$ & \multicolumn{2}{c}{$0.95$} \\
Numerical regularizer $\epsilon$ & \multicolumn{2}{c}{$10^{-8}$} \\
Correctness floor $\lambda_{\min}$ & \multicolumn{2}{c}{$0.3$} \\
Target floor $L_{\min}$ & \multicolumn{2}{c}{$1{,}000$} \\
Initial weights $\boldsymbol\lambda^{(0)}$ & \multicolumn{2}{c}{$(0.5, 0.5)$} \\
\bottomrule
\end{tabular}
\end{table}

\subsection{Training dynamics on DeepSeek-R1-Distill-Qwen-1.5B}
\label{app:dynamics}

\begin{figure}[t]
    \centering
    \includegraphics[width=0.65\linewidth]{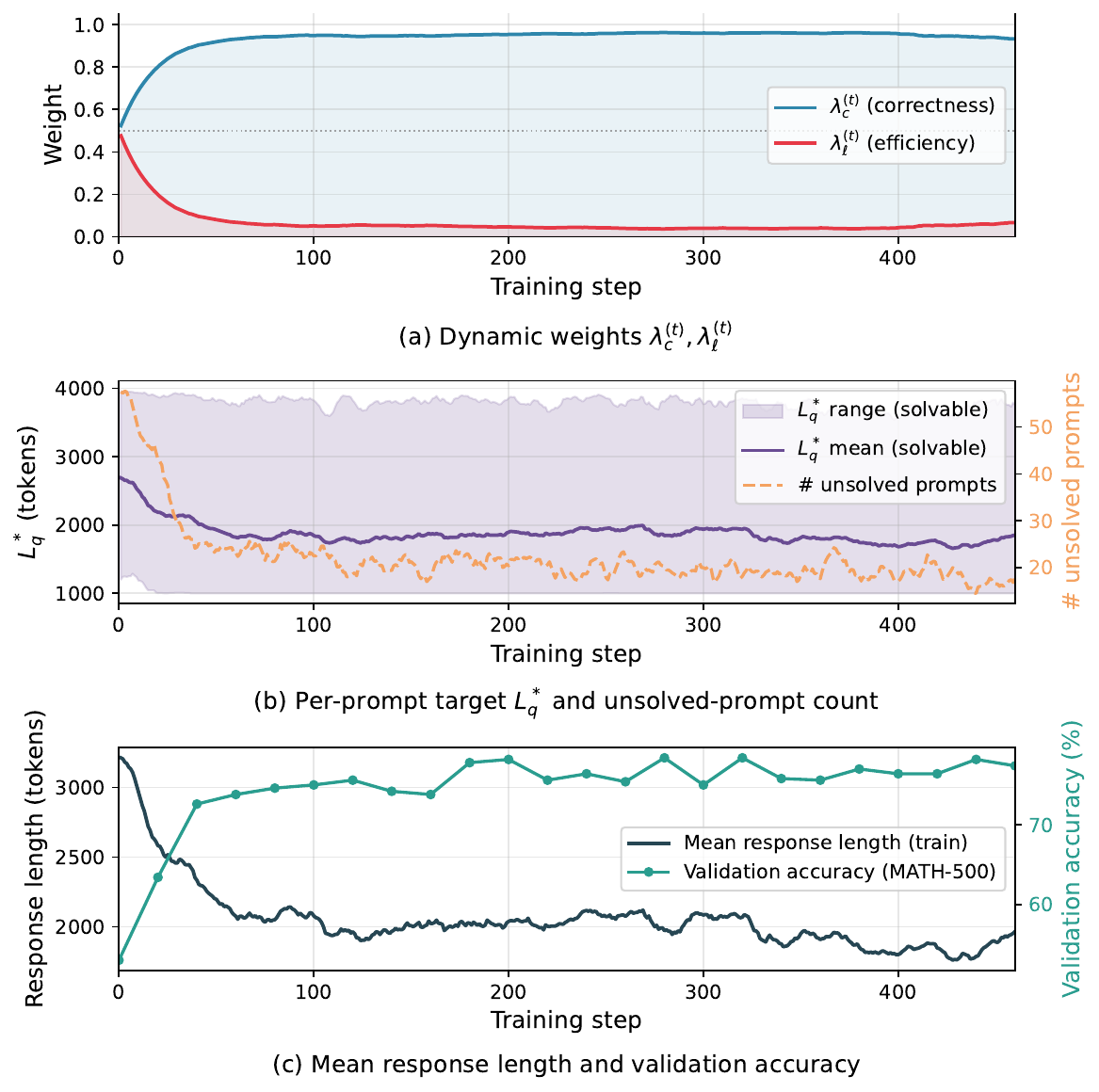}
    \caption{\textbf{Training dynamics of a LEAD run on DeepSeek-R1-Distill-Qwen-1.5B} (4K budget, $B_\mathrm{max}{=}4{,}000$).
    \textbf{(a)} Dynamic weights $\lambda_c^{(t)}, \lambda_\ell^{(t)}$.
    \textbf{(b)} Per-prompt $L^*_q$ statistics across solvable prompts (mean and min--max range) and the count of unsolved prompts (assigned $B_\mathrm{max}$).
    \textbf{(c)} Rolling mean response length on the rollout batch and validation accuracy on MATH-500.}
    \label{fig:dynamics}
\end{figure}

Figure~\ref{fig:dynamics} reports three signals over the course of a single LEAD run on DeepSeek-R1-Distill-Qwen-1.5B (290 logged steps): (a) the dynamic weights $\lambda_c^{(t)}, \lambda_\ell^{(t)}$, (b) the per-prompt $L^*_q$ statistics (mean and min--max range across solvable prompts, plus the count of unsolved prompts assigned the sentinel $B_\mathrm{max}$), and (c) the rolling mean response length and validation accuracy on MATH-500. Together they show that LEAD behaves as designed: the controller smoothly reweights the two objectives without manual scheduling, $L^*_q$ tightens online as the model improves, and length drops while accuracy keeps rising.

The trajectory of $\lambda_\ell$ in Figure~\ref{fig:dynamics}(a) reveals a key mechanism: \emph{the efficiency reward acts as a transient curriculum signal}. From the uniform initialization $(\lambda_c, \lambda_\ell){=}(0.5, 0.5)$ the controller shifts rapidly toward correctness in the early ``learn-to-solve'' phase, reaching $\lambda_\ell\!\approx\!0.08$ by step $50$ and plateauing near $\lambda_c\!\approx\!0.93$, $\lambda_\ell\!\approx\!0.07$ for the remainder of training. This early window is precisely when the bulk of length compression occurs in (c) and when $L^*_q$ tightens most rapidly in (b). Once a concise reasoning style is established and the efficiency reward saturates (its within-group CV collapses), the controller automatically holds gradient capacity on correctness, where signal still remains. Crucially, this is not equivalent to training without a length reward: if we had ablated the efficiency term entirely from the start, the model would never have learned the early length-compression behavior that makes the late-phase correctness-only optimization possible.

\subsection{Token allocation by prompt difficulty}
\label{app:difficulty}

\begin{figure}[t]
    \centering
    \includegraphics[width=0.95\linewidth]{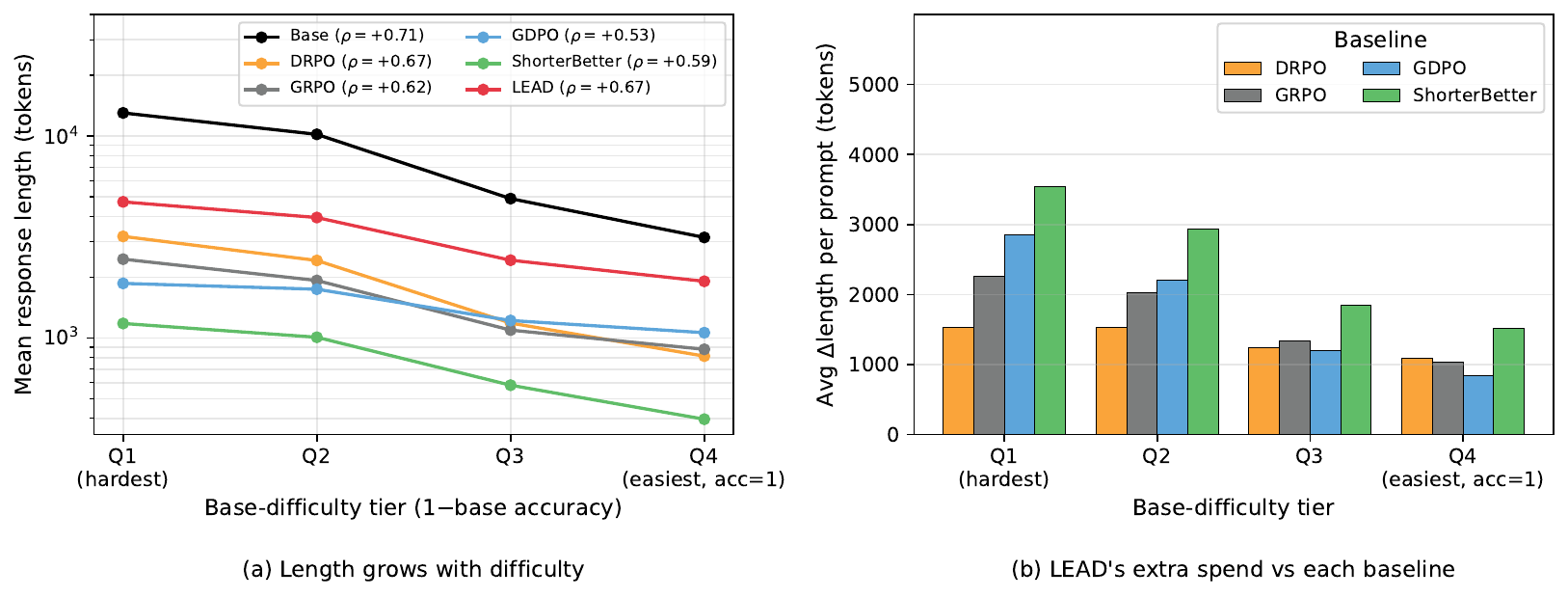}
    \caption{\textbf{Per-prompt token allocation by base-difficulty tier} on the pooled $5$-benchmark eval set ($1{,}275$ prompts). Difficulty is $1 - \mathrm{acc}_{\mathrm{base}}(q)$ from the unmodified base model. Prompts are grouped into four tiers by base pass-rate; $Q_4$ collapses all $\mathrm{acc}{=}1$ prompts ($510$ prompts) into one bin to avoid an arbitrary task-driven split of tied perfect-pass prompts.
    \textbf{(a)} Mean response length per tier, with Spearman $\rho$ between difficulty and length in the legend (higher $\rho$ = greater difficulty-sensitivity). LEAD ($\rho{=}+0.67$) tracks the base model ($\rho{=}+0.71$) most closely; GDPO ($\rho{=}+0.53$) compresses most uniformly.
    \textbf{(b)} Average per-prompt extra tokens spent by LEAD over each baseline, $\Delta\ell(q) = \ell_{\mathrm{LEAD}}(q) - \ell_{B}(q)$, averaged within each tier. Against the strongest compression baselines (GDPO, ShorterBetter, GRPO), the gap on the hardest tier is $2.2$--$3.4\times$ larger than on the easiest, showing that LEAD's extra tokens go preferentially to harder prompts. Note: this panel aggregates per-prompt across the $1{,}275$-prompt evaluation set (so MATH-500 and OlympiadBench, which are larger, dominate the overall mean), whereas Table~\ref{tab:main_results}'s ``Average Length'' is averaged across the five benchmarks with equal weight; the two aggregations can disagree on which method is shorter overall when per-task mean lengths differ widely.}
    \label{fig:difficulty}
\end{figure}

We probe whether LEAD allocates its token budget non-uniformly across prompts, as the per-problem $L^*_q$ design predicts. We define the difficulty of an evaluation prompt $q$ as $\mathrm{diff}(q) = 1 - \mathrm{acc}_{\mathrm{base}}(q)$, where $\mathrm{acc}_{\mathrm{base}}(q)$ is the avg@$n$ accuracy of the \emph{unmodified base} DeepSeek-R1-Distill-Qwen-1.5B on $q$ (using the same $n$ and decoding settings as the main evaluation: $n{=}10$ for AIME 2024/25 and AMC 2023, $n{=}3$ for MATH-500 and OlympiadBench, temperature $0.8$, top-$p$ $0.9$). Difficulty is defined externally via the base model so it does not depend on which method we are scoring. We pool the $1{,}275$ prompts across the five benchmarks, rank by base accuracy (with ties broken by first-appearance ordering), and partition into four ordered tiers: $Q_1$ (hardest, base pass-rate $0$, $255$ prompts) through $Q_3$, plus $Q_4$ (easiest, base pass-rate $1$, $510$ prompts). We collapse the perfect-pass tier into a single bin because avg@$n$ is discrete and roughly $40\%$ of prompts hit $\mathrm{acc}{=}1$, so a fifth quantile bin would split tied prompts arbitrarily by benchmark composition rather than by difficulty. Figure~\ref{fig:difficulty}(a) plots the mean response length per tier (log scale). The base model's strong positive dependence between difficulty and length (Spearman $\rho{=}+0.71$, computed per-prompt) reflects the natural fact that harder problems require more reasoning. LEAD preserves this structure ($\rho{=}+0.67$), staying closer to the base curve than any compression baseline; GDPO ($\rho{=}+0.53$) flattens the curve the most. Figure~\ref{fig:difficulty}(b) makes the allocation gap explicit: for each baseline $B$ we plot the average per-prompt $\Delta\ell(q) = \ell_{\mathrm{LEAD}}(q) - \ell_{B}(q)$ within each tier. The extra tokens LEAD spends are concentrated on the hardest prompts ($+1{,}540$ to $+3{,}540$ tokens on $Q_1$ vs.\ baselines, dropping to $+850$ to $+1{,}510$ on $Q_4$). The skew toward hard prompts is most pronounced against the strongest compression baselines (GDPO $3.4\times$, ShorterBetter $2.3\times$, GRPO $2.2\times$); against DRPO, which already preserves length on hard problems, LEAD's extra spend is closer to uniform ($1.4\times$). This is the mechanism behind LEAD's higher AES despite its longer average length: LEAD does not blanket-compress; it rations its budget by per-problem difficulty.

\paragraph{Computation of $\rho$.} The Spearman $\rho$ values shown in the legend of Figure~\ref{fig:difficulty}(a) are computed at \emph{prompt granularity} across the $1{,}275$-prompt evaluation set, not from the four tier means in the plot (the tiers are display-only). For each method $m$, we rank both the per-prompt difficulty vector $\mathrm{diff}(q)$ and the per-prompt mean-length vector $\bar{\ell}_m(q) = \tfrac{1}{n_q}\sum_j \ell_{m,q,j}$ (ties broken by first-appearance order), and take the Pearson correlation between the two rank vectors, which equals the standard Spearman rank correlation. $\rho$ is invariant under monotone rescaling of either axis, so it is unaffected by the log-axis in the plot. $\rho{=}+1$ means the method's per-prompt response length perfectly tracks base difficulty; $\rho{=}0$ indicates no monotonic rank association and does not by itself imply uniform response lengths.

\subsection{Results at 8K budget on DeepSeek-R1-Distill-Qwen-1.5B}
\label{app:1.5b_8k}

We additionally train and evaluate every method on DeepSeek-R1-Distill-Qwen-1.5B with the training-time max response length raised from 4K to 8K, keeping all other hyperparameters and the evaluation protocol identical to Section~\ref{sec:experimental_setup}. As shown in Table~\ref{tab:1.5b_8k_results}, the same conclusions as the 4K main results hold: LEAD again posts the highest accuracy ($54.44$, $+2.59$ over the base) and the highest AES ($0.54$). GDPO closes the AES gap to $0.01$ but still trails LEAD by $3.17$ accuracy points. ShorterBetter remains the weakest, with a $7.3$-point accuracy regression and the worst AES among baselines, as forcing every correct rollout toward a single minimum length still over-compresses on hard problems. Appendix~\ref{app:uncertainty} reports accuracy confidence intervals for these rows, and Appendix~\ref{app:checkpoint_audit} re-evaluates the public 8K checkpoint.

\begin{table}[t]
\centering
\caption{Performance comparison across methods on DeepSeek-R1-Distill-Qwen-1.5B at the 8K training-time max response length. AES denotes the Accuracy-Efficiency Score. \textbf{Bold} marks the best Acc / AES among trained methods and \underline{underline} marks the second best (the Base row is a reference and excluded from ranking).}
\label{tab:1.5b_8k_results}
\resizebox{\textwidth}{!}{%
\begin{tabular}{l cc cc cc cc cc | cc c}
\toprule
\multirow{2}{*}{\textbf{Model}} & \multicolumn{2}{c}{\textbf{AIME24}} & \multicolumn{2}{c}{\textbf{AIME25}} & \multicolumn{2}{c}{\textbf{AMC23}} & \multicolumn{2}{c}{\textbf{MATH}} & \multicolumn{2}{c|}{\textbf{OlyBch}} & \multicolumn{2}{c}{\textbf{Average}} & \textbf{AES~$\uparrow$} \\
\cmidrule(lr){2-3} \cmidrule(lr){4-5} \cmidrule(lr){6-7} \cmidrule(lr){8-9} \cmidrule(lr){10-11} \cmidrule(lr){12-13} \cmidrule(lr){14-14}
 & Acc & Len & Acc & Len & Acc & Len & Acc & Len & Acc & Len & Acc & Len &  \\
\midrule
Base            & 29.33 & 13018 & 24.00 & 13126 & 69.25 & 7472 & 84.93 & 4080 & 51.75 & 8370 & 51.85 & 9213 & -- \\
\midrule
GRPO            & 22.67 & 4275 & 20.00 & 3766 & \underline{71.50} & 2607 & 84.80 & 1572 & 51.16 & 2813 & 50.03 & 3007 &  0.32 \\
GDPO            & \underline{29.33} & 4736 & 21.33 & 4058 & 67.00 & 2795 & \underline{85.20} & 1665 & \underline{53.48} & 3005 & 51.27 & 3252 & \underline{0.53} \\
ShorterBetter   & 22.67 & 4305 & 14.67 & 3934 & 62.00 & 1948 & 78.93 & 1109 & 44.69 & 2271 & 44.59 & 2713 & $-$0.69 \\
DRPO            & 28.00 & 4472 & \underline{22.33} & 4779 & 70.00 & 3665 & 83.13 & 6005 & 53.23 & 4255 & \underline{51.34} & 4635 &  0.40 \\
\textbf{LEAD (Ours)} & \textbf{30.33} & 8219 & \textbf{26.00} & 7874 & \textbf{73.50} & 4368 & \textbf{86.53} & 2624 & \textbf{55.85} & 5015 & \textbf{54.44} & 5620 & \textbf{0.54} \\
\bottomrule
\end{tabular}%
}
\end{table}

\section{Additional protocol details and robustness evidence}
\label{app:additional_evidence}

\subsection{Evaluation and training-comparison conventions}
\label{app:protocol_detail}

\paragraph{Accuracy and response length.}
For benchmark $b$ with $N_b$ questions, let $y_{bi s}\in\{0,1\}$ denote correctness of decode $s$ for question $i$. We compute
\begin{equation}
 a_{bi}=\frac{1}{n_b}\sum_{s=1}^{n_b}y_{bi s},\qquad
 A_b=\frac{100}{N_b}\sum_{i=1}^{N_b}a_{bi},\qquad
 A=\frac{1}{5}\sum_{b=1}^{5}A_b.
\end{equation}
This is avg@$n$, not pass@$n$. The five tasks are AIME24, AIME25, AMC23, MATH-500, and OlympiadBench. Minerva outputs generated by the shared harness are excluded from the paper aggregate. Length is averaged across all scored response texts within a benchmark, then equally across benchmarks. The aggregation script re-tokenizes response text using the DeepSeek-R1-Distill-Qwen-1.5B tokenizer with special-token insertion disabled. It excludes the prompt and includes incorrect responses. This text-token count is distinct from the training response-mask length.

\paragraph{Evaluation budget and scoring.}
The frozen evaluator uses temperature $0.8$, top-$p=0.9$, seed 0, and nominal settings of 32,768 for both maximum new tokens and model context length. The context budget includes prompt tokens, which reduce the available completion budget. Generation also stops at the model's termination condition. Responses reaching the length limit are retained and graded from their emitted text using the same answer-extraction and verification pipeline. We retain the runtime configuration, tokenizer, prompt template, and generated outputs for reproducibility.

\paragraph{Training configurations.}
The original training comparisons share data, base models, and nominal training-response caps, with method-specific reward and optimization settings. The 1.5B scalar-reward GRPO sweep uses learning rate $10^{-6}$, upper PPO clip parameter $0.28$, group filtering, and a fixed-budget indicator reward. Dynamic LEAD uses learning rate $2\times10^{-6}$, upper clip $0.2$, no group filtering, and the symmetric target reward. The matched Linear decay schedule/PSI pair uses the same data, seed 42, four-L40S topology, learning rate $2\times10^{-6}$, clip parameters $0.2/0.2$, no group filtering, $128$ prompts, $G=8$, 4,000-token cap, and seven-epoch training configuration. Only the channel-weight controller changes in this pair.

\subsection{Data-order-seed variation and question uncertainty}
\label{app:uncertainty}

\paragraph{Three-run summaries.}
Table~\ref{tab:seed_summary} summarizes the three runs and Table~\ref{tab:seed_individual} gives the individual runs.
\begin{table}[htbp]
\centering
\caption{Results over three data-order seeds on DeepSeek-R1-Distill-Qwen-1.5B with a 4K training cap, reported as mean $\pm$ sample SD. \textbf{Bold} marks the best mean Acc / AES and \underline{underline} marks the second best.}
\label{tab:seed_summary}
\small
\begin{tabular}{lrrr}
\toprule
Method & Avg Acc & Avg Len & AES \\
\midrule
GRPO & $47.82\pm1.25$ & $2078\pm219$ & $0.00\pm0.24$ \\
GDPO & $46.25\pm1.26$ & $1598\pm195$ & $-0.25\pm0.23$ \\
DRPO & $\underline{50.82\pm0.67}$ & $2752\pm124$ & $\underline{0.50\pm0.12}$ \\
\textbf{LEAD} & $\mathbf{52.93\pm0.91}$ & $3589\pm108$ & $\mathbf{0.67\pm0.05}$ \\
\bottomrule
\end{tabular}
\end{table} The varied seed controls training data order. Seed-42 rows retain their original precision, and new aggregates use the five per-benchmark results before final table rounding. AES is computed per run against $51.85/9213$ and then averaged. GRPO's mean AES is slightly negative before rounding ($-0.002$).

\begin{table}[htbp]
\centering\small
\caption{Individual 1.5B/4K data-order-seed runs. Each cell is Avg Acc / Avg Len / AES.}
\label{tab:seed_individual}
\begin{tabular}{lrrr}
\toprule
Method & Seed 42 (submitted) & Seed 43 & Seed 44 \\
\midrule
GRPO & 48.12 / 1847 / 0.080 & 48.89 / 2282 / 0.182 & 46.46 / 2104 / $-0.269$ \\
GDPO & 45.49 / 1643 / $-0.405$ & 45.56 / 1385 / $-0.364$ & 47.71 / 1767 / 0.009 \\
DRPO & 50.74 / 2647 / 0.499 & 51.53 / 2889 / 0.624 & 50.19 / 2719 / 0.384 \\
LEAD & 53.36 / 3714 / 0.684 & 51.88 / 3526 / 0.619 & 53.55 / 3528 / 0.715 \\
\bottomrule
\end{tabular}
\end{table}

We use sample SD with denominator $n-1$ and the Student-$t$ interval $\bar{x}\pm t_{0.975,2}s/\sqrt{3}$, where $t_{0.975,2}=4.303$. Table~\ref{tab:seed_ci} reports the resulting 95\% confidence intervals. With three runs, these intervals are wide and depend on the Student-$t$ assumptions. LEAD's mean accuracy exceeds DRPO's by $2.11$ points.

\begin{table}[htbp]
\centering\small
\caption{Across-run 95\% Student-$t$ confidence intervals for the means of the three data-order-seed runs. These are distinct from question-bootstrap intervals.}
\label{tab:seed_ci}
\begin{tabular}{lrrr}
\toprule
Method & Avg Acc & Avg Len & AES \\
\midrule
GRPO & [44.73, 50.92] & [1534, 2621] & [$-0.589$, 0.585] \\
GDPO & [43.12, 49.38] & [1115, 2082] & [$-0.820$, 0.313] \\
DRPO & [49.14, 52.49] & [2443, 3060] & [0.204, 0.801] \\
LEAD & [50.66, 55.20] & [3321, 3857] & [0.551, 0.795] \\
\bottomrule
\end{tabular}
\end{table}

\paragraph{Paired question bootstrap.}
For each comparison, we resample questions separately within each benchmark, retaining the complete avg@$n$ score of each sampled question. We use the same sampled question indices for the two methods, compute five benchmark differences, and average them equally. The 2.5th and 97.5th percentiles of 10,000 resamples (seed 0) give Table~\ref{tab:paired_ci}. These intervals quantify question-resampling uncertainty for the evaluated models and sampled decodes, complementing the training-run summaries above. The 7B difference is rounded from per-question results and can differ by 0.01 from subtracting displayed table cells.

\begin{table}[htbp]
\centering\small
\caption{Paired 95\% question-bootstrap intervals for average-accuracy differences, in percentage points.}
\label{tab:paired_ci}
\begin{tabular}{lrr}
\toprule
Comparison & Difference & 95\% CI \\
\midrule
1.5B/4K: LEAD $-$ DRPO & $+2.62$ & [$+0.79$, $+4.52$] \\
1.5B/8K: LEAD $-$ GDPO & $+3.17$ & [$+1.31$, $+5.33$] \\
7B/4K: LEAD $-$ DRPO & $+1.17$ & [$-0.65$, $+3.09$] \\
\bottomrule
\end{tabular}
\end{table}
Both 1.5B intervals exclude zero, supporting LEAD's accuracy gains over these baselines under question resampling. The 7B interval includes zero.

Table~\ref{tab:all_question_ci} provides the marginal accuracy intervals for the original 4K and 8K comparisons. These accompany the submitted point estimates. Recomputing from per-question scores can change a displayed average by 0.01 through rounding.
\begin{table}[htbp]
\centering
\caption{Marginal 95\% question-bootstrap intervals for accuracy of the original models (10,000 resamples). The average gives equal weight to the five benchmarks. The intervals quantify question-resampling uncertainty for these fixed models.}
\label{tab:all_question_ci}
\resizebox{\textwidth}{!}{\begin{tabular}{lrrrrrr}
\toprule
Setting / method & AIME24 & AIME25 & AMC23 & MATH-500 & OlympiadBench & Avg Acc \\
\midrule
1.5B-4K Base & [18.00, 41.33] & [12.00, 37.33] & [59.75, 78.75] & [82.40, 87.47] & [48.49, 55.01] & [47.81, 55.98] \\
1.5B-4K GRPO & [14.00, 34.67] & [8.67, 31.67] & [55.49, 75.75] & [79.47, 85.00] & [45.88, 52.44] & [44.35, 52.01] \\
1.5B-4K GDPO & [9.67, 31.67] & [6.33, 28.33] & [49.75, 72.00] & [77.80, 83.73] & [45.38, 52.00] & [41.65, 49.46] \\
1.5B-4K ShorterBetter & [14.67, 35.33] & [8.67, 30.33] & [54.75, 75.00] & [77.53, 83.20] & [41.63, 47.95] & [43.04, 50.46] \\
1.5B-4K DRPO & [15.67, 40.00] & [10.00, 33.67] & [60.00, 80.25] & [80.53, 86.00] & [48.49, 55.11] & [46.76, 54.86] \\
1.5B-4K LEAD & [22.67, 48.33] & [12.33, 37.67] & [56.00, 78.00] & [82.87, 88.00] & [51.16, 57.78] & [49.14, 57.71] \\
\midrule
7B-4K Base & [39.66, 67.67] & [26.00, 52.33] & [80.00, 94.50] & [91.73, 95.27] & [63.90, 70.12] & [63.93, 72.37] \\
7B-4K GRPO & [25.67, 50.33] & [15.67, 40.33] & [73.75, 91.51] & [86.60, 91.13] & [55.70, 62.17] & [55.34, 63.32] \\
7B-4K GDPO & [27.33, 56.00] & [14.67, 41.67] & [70.25, 89.25] & [87.33, 91.93] & [56.64, 63.36] & [55.40, 64.38] \\
7B-4K ShorterBetter & [33.67, 59.00] & [18.33, 45.00] & [72.25, 88.25] & [80.87, 86.13] & [57.33, 63.70] & [56.41, 64.68] \\
7B-4K DRPO & [30.00, 58.67] & [17.67, 44.67] & [80.50, 94.25] & [90.00, 94.07] & [61.88, 68.44] & [59.81, 68.24] \\
7B-4K LEAD & [31.67, 58.00] & [22.00, 48.67] & [80.50, 95.75] & [90.33, 94.13] & [61.83, 68.30] & [61.09, 69.31] \\
\midrule
1.5B-8K LEAD & [18.67, 43.00] & [13.67, 39.33] & [63.25, 83.00] & [83.87, 89.00] & [52.49, 59.16] & [50.25, 58.74] \\
1.5B-8K GDPO & [18.33, 41.33] & [9.67, 34.33] & [56.25, 77.25] & [82.67, 87.67] & [50.12, 56.79] & [47.26, 55.32] \\
1.5B-8K GRPO & [12.00, 34.67] & [9.00, 32.33] & [61.00, 81.25] & [82.13, 87.40] & [47.75, 54.47] & [46.21, 54.10] \\
1.5B-8K DRPO & [17.00, 42.67] & [11.33, 35.33] & [61.75, 81.00] & [79.07, 84.33] & [47.56, 53.98] & [47.23, 55.45] \\
\bottomrule
\end{tabular}}
\end{table}

\subsection{Matched schedule comparisons}
\label{app:schedules}
The Linear decay schedule sets $\lambda_\ell(t)=0.5-0.45\min((t-1)/49,1)$ for training updates $t\ge1$ and $\lambda_c(t)=1-\lambda_\ell(t)$: the efficiency weight is $0.5$ at update 1, reaches $0.05$ at update 50, and stays there. All remaining target, reward, normalization, and optimizer settings match PSI. The Two-phase schedule keeps equal weights for the first 50 updates and switches to correctness only afterward; its run uses learning rate $10^{-6}$, upper clip $0.28$, and group filtering. Table~\ref{tab:schedule_matched} reports the three controllers. Linear decay is competitive with PSI at step 300, PSI improves both accuracy and length relative to it at step 450, and the Two-phase schedule is below both.

\begin{table}[htbp]
\centering\small
\caption{Weighting-controller comparison on DeepSeek-R1-Distill-Qwen-1.5B (4K), seed 42. This PSI run is distinct from the LEAD run in Table~\ref{tab:main_results}.}
\label{tab:schedule_matched}
\begin{tabular}{llrrr}
\toprule
Step & Controller & Avg Acc & Avg Len & AES \\
\midrule
300 & Linear decay schedule & 53.97 & 3640 & 0.727 \\
300 & PSI & 53.53 & 3561 & 0.710 \\
\midrule
450 & Linear decay schedule & 52.29 & 3637 & 0.631 \\
450 & Two-phase schedule & 50.76 & 3169 & 0.446 \\
450 & PSI & 53.89 & 3343 & 0.755 \\
\bottomrule
\end{tabular}
\end{table}

\subsection{Released LASER-family checkpoint comparison}
\label{app:laser_comparison}
Table~\ref{tab:laser_all} compares LEAD with all nine released 1.5B LASER/LASER-D/LASER-DE checkpoints. Table~\ref{tab:laser_bench} provides per-benchmark results at 1.5B and for two additional 7B adaptive checkpoints. The numeric suffix denotes the released checkpoint's target parameter. All checkpoints use our shared evaluation protocol. Their training protocols differ: LASER uses DeepScaleR (approximately 40K problems) and a 16K training-response cap, while LEAD uses 8,521 MATH Level 3 to 5 problems and a 4K cap.

\begin{table}[htbp]
\centering\small
\caption{Comparison with released LASER models on DeepSeek-R1-Distill-Qwen-1.5B. AES uses the base accuracy and length of $51.85/9213$. \textbf{Bold} marks the best Acc / AES and \underline{underline} marks the second best.}
\label{tab:laser_all}
\begin{tabular}{lrrr}
\toprule
Method & Avg Acc & Avg Len & AES \\
\midrule
LASER-2048 & 49.83 & 4459 & 0.13 \\
LASER-4096 & 51.93 & 5716 & 0.38 \\
LASER-8192 & 51.92 & 5898 & 0.36 \\
LASER-D-1024 & 49.39 & 4256 & 0.06 \\
LASER-D-2048 & 52.05 & 4427 & \underline{0.53} \\
LASER-D-4096 & 52.06 & 5009 & 0.47 \\
LASER-DE-1024 & 49.74 & 4187 & 0.14 \\
LASER-DE-2048 & 51.05 & 4479 & 0.36 \\
LASER-DE-4096 & \underline{52.07} & 5319 & 0.44 \\
\textbf{LEAD} & \textbf{53.36} & 3714 & \textbf{0.68} \\
\bottomrule
\end{tabular}
\end{table}

At 1.5B, LEAD is more accurate and shorter on aggregate than each of the nine evaluated checkpoints. Against LASER-D-2048, the highest-AES LASER checkpoint, LEAD gains $1.31$ accuracy points and uses approximately $713$ fewer tokens ($16.1\%$ shorter). All lengths are measured under our shared decoding and scoring configuration.

\paragraph{Additional 7B comparison.}
LASER-D-4096 obtains $68.71 / 3989 / 0.42$ (Acc / Len / AES), and LASER-DE-4096 obtains $68.19 / 3780 / 0.43$, compared with LEAD's $65.17 / 4417 / -0.11$. Both LASER variants are more accurate and shorter at this scale under the shared evaluator. AES uses the 7B base accuracy and length of $68.20/6631$, and the training-protocol differences described above apply at both scales.

\begin{table}[htbp]
\centering
\caption{Per-benchmark released LASER results. Each benchmark cell is Acc / Len. LEAD's corresponding values are in Table~\ref{tab:main_results}.}
\label{tab:laser_bench}
\resizebox{\textwidth}{!}{\begin{tabular}{llrrrrr}
\toprule
Scale & Checkpoint & AIME24 & AIME25 & AMC23 & MATH-500 & OlympiadBench \\
\midrule
1.5B & LASER-2048 & 26.67 / 6657 & 19.00 / 6235 & 68.00 / 3515 & 82.80 / 1923 & 52.69 / 3962 \\
 & LASER-4096 & 31.33 / 8476 & 21.33 / 7935 & 71.50 / 4598 & 84.07 / 2541 & 51.41 / 5029 \\
 & LASER-8192 & 31.33 / 7887 & 23.00 / 7753 & 71.25 / 5136 & 83.67 / 3143 & 50.37 / 5570 \\
 & LASER-D-1024 & 26.67 / 6506 & 20.33 / 5876 & 68.00 / 3452 & 82.00 / 1759 & 49.93 / 3685 \\
 & LASER-D-2048 & 30.33 / 6564 & 23.33 / 6161 & 73.00 / 3469 & 83.07 / 1969 & 50.52 / 3975 \\
 & LASER-D-4096 & 30.67 / 7293 & 20.67 / 6888 & 73.00 / 3993 & 83.40 / 2338 & 52.54 / 4532 \\
 & LASER-DE-1024 & 30.00 / 6205 & 20.00 / 5713 & 68.50 / 3430 & 82.07 / 1822 & 48.15 / 3765 \\
 & LASER-DE-2048 & 31.00 / 6674 & 20.67 / 6369 & 68.50 / 3517 & 83.87 / 1868 & 51.21 / 3965 \\
 & LASER-DE-4096 & 32.00 / 7751 & 21.33 / 7388 & 72.25 / 4162 & 82.13 / 2462 & 52.64 / 4833 \\
\midrule
7B & LASER-D-4096 & 53.67 / 5840 & 37.67 / 6306 & 90.75 / 2772 & 93.60 / 1622 & 67.85 / 3405 \\
 & LASER-DE-4096 & 50.00 / 5506 & 39.00 / 5833 & 90.75 / 2515 & 93.60 / 1612 & 67.60 / 3436 \\
\bottomrule
\end{tabular}}
\end{table}

\subsection{Qwen3-4B LoRA transfer results}
\label{app:qwen3_lora}
Table~\ref{tab:qwen3_lora_summary} summarizes the Qwen3-4B LoRA comparison and Table~\ref{tab:qwen3_lora_bench} reports per-benchmark results. The base and both trained methods use the same evaluation configuration, with $n=10$ sampled decodes per question for AIME24, AIME25, and AMC23, and $n=3$ for MATH-500 and OlympiadBench. Aggregate accuracy is computed from unrounded question-level scores with equal benchmark weighting.

\begin{table}[htbp]
\centering\small
\caption{Qwen3-4B LoRA transfer study. Acc and Len average the five benchmarks equally. AES uses the unmodified Qwen3-4B base. Bold marks the best Acc / AES among the trained methods.}
\label{tab:qwen3_lora_summary}
\begin{tabular}{lrrr}
\toprule
Method & Avg Acc & Avg Len & AES \\
\midrule
Qwen3-4B base & 78.90 & 9945 & 0.00 \\
\midrule
GRPO-LoRA & 70.95 & 4776 & $-0.49$ \\
\textbf{LEAD-LoRA} & \textbf{74.00} & 6908 & $\mathbf{-0.32}$ \\
\bottomrule
\end{tabular}
\end{table}

Both arms use Qwen3-4B with thinking enabled, the same 8,521 MATH training problems, data-order seed 42, 128 prompts per batch, and $G=8$ rollouts per prompt. The prompt and response caps are 2,048 and 4,000 tokens. LoRA uses rank 64, scaling parameter 32, and all linear layers. The learning rate is $3\times10^{-6}$, constant without warmup. Both arms use PPO mini-batches of 16 prompts, one PPO epoch, clipping thresholds of $0.2$, KL coefficient $0.0005$, and entropy coefficient $0.001$. The GRPO scalar length reward is a binary threshold at 2,000 character-estimated tokens, computed as the combined prompt and response character count divided by four. LEAD uses its symmetric response-token reward with $L_{\min}=1000$, $B_{\max}=4000$, $\beta=0.95$, and correctness-weight floor $0.3$.

Training uses verl v0.7.0 with the LEAD port and vLLM 0.11.0. LEAD runs on four L40S GPUs and GRPO on two H200 GPUs, with identical global batch and optimizer settings. All reported models are evaluated on an H200 using the same evaluation environment and decoding configuration. The confidence intervals use 10,000 paired question-bootstrap resamples and quantify question-sampling uncertainty rather than training-run variability.

For Table~\ref{tab:qwen3_lora_summary}, the 95\% accuracy intervals are $[75.01,82.67]$ for the base, $[66.68,75.11]$ for GRPO, and $[69.77,78.01]$ for LEAD. The paired LEAD-minus-GRPO difference is $+3.06$ points ($[+1.04,+5.13]$).

\begin{table}[htbp]
\centering
\caption{Per-benchmark Qwen3-4B LoRA results. Each benchmark cell reports Acc / Len.}
\label{tab:qwen3_lora_bench}
\resizebox{\textwidth}{!}{\begin{tabular}{lrrrrr}
\toprule
Method & AIME24 & AIME25 & AMC23 & MATH-500 & OlympiadBench \\
\midrule
Qwen3-4B base & 71.00 / 13189 & 61.33 / 15902 & 93.50 / 7000 & 95.40 / 4207 & 73.28 / 9428 \\
\midrule
GRPO-LoRA & 59.33 / 7083 & 44.00 / 7852 & 89.75 / 3072 & 93.20 / 1832 & 68.44 / 4040 \\
LEAD-LoRA & 66.00 / 10147 & 48.67 / 11315 & 90.50 / 4304 & 93.73 / 2741 & 71.11 / 6035 \\
\bottomrule
\end{tabular}}
\end{table}

\subsection{Public-checkpoint re-evaluation}
\label{app:checkpoint_audit}
The public 8K LEAD checkpoint, \texttt{Kotom1/math\_lead\_8k\_deepseek-r1-1.5b} at revision \texttt{f515d452}, obtains $54.93/5824/0.546$ under the frozen evaluator, compared with the original Table~\ref{tab:1.5b_8k_results} row $54.44/5620/0.54$. Its per-benchmark accuracies are $34.33,23.33,73.25,88.13,55.60$ in table order. We report this checkpoint re-evaluation separately from the original result. Re-evaluating the identical 4K checkpoint gives $53.16$ accuracy compared with the original $53.36$, illustrating variation in sampled evaluation across executions.

\subsection{BatchWhiten and the correctness-only comparison}
\label{app:whiten_detail}
Let $m_{qjt}\in\{0,1\}$ mask valid response-token positions and let $N=\sum_{qjt}m_{qjt}$. Broadcasting the rollout advantage $\tilde A_{q,j}$ across its response tokens, the implementation computes
\begin{equation}
 \bar\mu=\frac{1}{N}\sum_{qjt}m_{qjt}\tilde A_{q,j},\qquad
 \bar v=\frac{1}{N-1}\sum_{qjt}m_{qjt}(\tilde A_{q,j}-\bar\mu)^2,
\end{equation}
and returns $m_{qjt}(\tilde A_{q,j}-\bar\mu)/\sqrt{\bar v+10^{-8}}$. The training batches have more than one valid token. The utility rejects masks with zero or one valid position. Group normalization uses a sample standard deviation plus $\epsilon$, whereas PSI's pooled statistics use population variances.

At zero efficiency weight, reward-wise and scalar normalization coincide for correctness, while static LEAD additionally applies final BatchWhiten. All-equal correctness groups have zero correctness advantage before this operation. Their prevalence and response lengths affect the token-weighted batch moments. Whitening removes the batch offset and standardizes the surviving correctness contrasts, changing their scale relative to other loss terms while leaving the PPO clipping threshold unchanged. In a matched first-batch diagnostic, advantage mean/variance and gradient norm are $-0.048/0.439/0.628$ without final whitening and approximately $0/1/0.942$ with it. This additional normalization step is part of the advantage-processing difference in Table~\ref{tab:abl_dynamic}.

\subsection{PSI edge cases and usable ranking contrasts}
\label{app:psi_edges}
For correctness, near-failure has $\mu_c\approx0$ and $P_c\approx1$, whereas near-saturation has $\mu_c\approx1$ and $P_c\approx0$. The headroom factor distinguishes these reward means even when both variances are small. Within an all-wrong or all-correct prompt, correctness advantages are zero because all rewards are equal. The implementation adds $10^{-8}$ before taking the controller standard deviation, providing numerical stabilization even for constant rewards. If the whole batch has no correct rollout, the efficiency controller moments are set to zero. If both PSI scores vanish, their stabilized target ratios are zero and the pre-floor EMA decays. The correctness floor and complementary efficiency weight determine the applied pair. Fixed reward bounds keep the potential denominator nonzero.

\subsection{Sparse-correct targets and the zero-correct fallback}
\label{app:sparse_targets}
Table~\ref{tab:sparse_hist} reports correct-count histograms from the matched seed-42 PSI and \emph{Linear decay schedule} runs. These additional runs share the target construction and $128\times8$ rollout geometry. We use identical windows, with 128 prompt groups per update. The PSI run resumed from step 100, and repeated steps 101 to 113 use the resumed trajectory. Groups with one or two correct rollouts are a minority in both runs and become less frequent as training progresses.

\begin{table}[htbp]
\centering\small
\caption{Correct-count distributions in the additional matched seed-42 runs. Percentages are computed over prompt groups and can differ from 100\% by rounding. The last row for each controller is a 100-update late-training window.}
\label{tab:sparse_hist}
\begin{tabular}{llrrrr}
\toprule
Controller & Steps & $|\mathcal C_q|=0$ & 1-2 & 3-5 & 6-8 \\
\midrule
PSI & 1-50 & 13.33\% & 8.95\% & 11.94\% & 65.78\% \\
 & 51-300 & 10.05\% & 6.65\% & 10.96\% & 72.34\% \\
 & 301-460 & 8.62\% & 4.62\% & 7.49\% & 79.27\% \\
 & 361-460 & 8.34\% & 4.41\% & 7.16\% & 80.09\% \\
\midrule
Linear decay & 1-50 & 13.84\% & 7.80\% & 13.36\% & 65.00\% \\
 & 51-300 & 8.81\% & 5.61\% & 9.59\% & 76.00\% \\
 & 301-460 & 7.39\% & 4.59\% & 7.56\% & 80.46\% \\
 & 361-460 & 7.06\% & 4.42\% & 7.39\% & 81.13\% \\
\bottomrule
\end{tabular}
\end{table}

The target lies in $[1000,B_{\max}]$, raw efficiency rewards lie in $[-1,1]$, and advantage normalization uses all eight rollout rewards with numerical stabilization. These safeguards keep raw values bounded and provide a full group for normalization. Correct rollouts supply positive-versus-negative correctness contrast and examples of successful reasoning lengths. With one correct sample, mean, median, and minimum coincide. With two, the conventional median equals the mean. The aggregator comparison in Table~\ref{tab:abl_aggregator} therefore reflects differences primarily in larger correct sets. Sparse-sample target uncertainty remains a limitation discussed in Appendix~\ref{app:limitations_impact}.

A separate control sets $r_\ell=0$ for every rollout when $|\mathcal C_q|=0$. It obtains $52.06/3719/0.594$, compared with $52.84/3551/0.659$ for the corresponding full-fallback run. These runs are separate from the main comparison. Substantial accuracy and length gains remain after removing the fallback, with a $0.78$-point accuracy difference between the two runs.

\subsection{Observed target movement and the training cap}
\label{app:target_dynamics}
Table~\ref{tab:target_windows} uses the submitted PSI training log on MATH Level 3-5. Target means and medians are computed over solvable prompt groups and then averaged across updates in each window. Accuracy and cap-hit rate use all rollouts. Both the mean and median target decrease as training-batch accuracy rises.

\begin{table}[htbp]
\centering\small
\caption{Submitted 1.5B/4K PSI trajectory: window averages. Target statistics exclude zero-correct groups.}
\label{tab:target_windows}
\begin{tabular}{lrrrr}
\toprule
Steps & Mean $L_q^*$ & Median $L_q^*$ & Batch accuracy & Cap-hit rate \\
\midrule
1-10 & 2628 & 2619 & 0.427 & 0.471 \\
50-59 & 1868 & 1681 & 0.680 & 0.154 \\
451-460 & 1830 & 1677 & 0.780 & 0.050 \\
\bottomrule
\end{tabular}
\end{table}

The cap bounds the available rollout lengths, while correctness filtering selects the trajectories used to re-estimate the target. Longer incomplete or incorrect rollouts are excluded from this correct set. Selection and re-estimation can therefore contribute to downward target movement as training progresses. The symmetric reward encourages lengths near the current target, with re-estimation allowing that preferred length to change.

Table~\ref{tab:abl_dynamic}'s correctness-only static LEAD reaches $50.60/3141/0.42$, compared with dynamic LEAD's $53.36/3714/0.68$. Compression is therefore already present without the efficiency channel, while the full method preserves more reasoning tokens and achieves higher accuracy and AES. The 8K results further show that LEAD retains compression relative to the base under a relaxed cap. Together, these experiments characterize the accuracy and length trade-off under different reward settings and training budgets.

\subsection{Behavioral-floor diagnostic}
\label{app:behavioral_floor}
We analyze per-prompt correct-rollout lengths from the \emph{neutral-fallback PSI control}, which uses LEAD's solved-prompt target and sets the zero-correct efficiency reward to zero. Over its final 100 updates (362 to 461), 8,877 of 12,800 prompt groups have $|\mathcal C_q|\ge4$. Among these eligible groups, the median within-correct length CV is $0.163$, and the shortest observed correct rollout is a median $22.4\%$ shorter than its group mean. Only $3.85\%$ have a gap of at most $10\%$.

To measure how often correct lengths cluster tightly at a high value, we count groups with mean correct length at least $b$ and within-correct population CV at most $\tau$. Table~\ref{tab:floor_sensitivity} reports sensitivity to both thresholds. At $b=2000$ and $\tau=0.10$, 223 of 8,877 eligible groups meet the criterion ($2.51\%$, or $1.74\%$ of all groups). Tight observed clusters are uncommon at these thresholds. These measurements describe sampled trajectories, while the minimum sufficient reasoning length remains unknown.

\begin{table}[htbp]
\centering\small
\caption{High-and-tight proxy frequency among eligible groups ($|\mathcal C_q|\ge4$) in the neutral-fallback PSI control, final 100 updates. The thresholds $b$ and $\tau$ define the observed clustering criterion.}
\label{tab:floor_sensitivity}
\begin{tabular}{lrrrr}
\toprule
Mean correct length $b$ & $\tau=0.05$ & $\tau=0.10$ & $\tau=0.15$ & $\tau=0.20$ \\
\midrule
1500 & 0.35\% & 4.66\% & 15.51\% & 26.79\% \\
2000 & 0.27\% & 2.51\% & 7.25\% & 12.40\% \\
2500 & 0.20\% & 1.17\% & 3.18\% & 4.88\% \\
\bottomrule
\end{tabular}
\end{table}

\newpage
\section*{NeurIPS Paper Checklist}

\begin{enumerate}

\item {\bf Claims}
    \item[] Question: Do the main claims made in the abstract and introduction accurately reflect the paper's contributions and scope?
    \item[] Answer: \answerYes{} 
    \item[] Justification: The abstract and Section~\ref{introduction} state two contributions: (i) PSI-driven dynamic reward weighting with decoupled group normalization, and (ii) per-problem online length target $L^*_q$ with a symmetric efficiency reward. Both are formalized in Section~\ref{method} and validated empirically in Section~\ref{experiment}.

\item {\bf Limitations}
    \item[] Question: Does the paper discuss the limitations of the work performed by the authors?
    \item[] Answer: \answerYes{} 
    \item[] Justification: A dedicated \emph{Limitations and Broader Impact} section in Appendix~\ref{app:limitations_impact} discusses four scope-defining aspects of LEAD: (i) it targets RL settings with reliably verifiable correctness and group sampling, with extensions to open-ended generation requiring task-specific reward models, (ii) the bootstrapping design intentionally prioritizes correctness before compression on unsolved prompts, (iii) the single-scalar per-problem target fits unimodal-length reasoning tasks but extending to distributional or multi-target length models is a future direction, (iv) LEAD is a training-time mechanism rather than a deployment-time budget controller, and combining it with test-time budget allocation is left as future work.

\item {\bf Theory assumptions and proofs}
    \item[] Question: For each theoretical result, does the paper provide the full set of assumptions and a complete (and correct) proof?
    \item[] Answer: \answerNA{}
    \item[] Justification: The paper presents an algorithmic/empirical contribution. It does not contain formal theorems requiring proofs.

    \item {\bf Experimental result reproducibility}
    \item[] Question: Does the paper fully disclose all the information needed to reproduce the main experimental results of the paper to the extent that it affects the main claims and/or conclusions of the paper (regardless of whether the code and data are provided or not)?
    \item[] Answer: \answerYes{}
    \item[] Justification: Section~\ref{sec:experimental_setup} specifies the dataset (MATH level 3-5), base models (DeepSeek-R1-Distill-Qwen-1.5B/7B), all training hyperparameters (learning rate, batch sizes, KL coefficient, rollout count, epochs), and the LEAD-specific hyperparameters ($\alpha, \beta, \lambda_{\min}, B_{\max}$). Algorithm~\ref{alg:lead} in Appendix~\ref{app:algorithm} gives the full advantage-computation procedure. The evaluation protocol follows Sober Reasoning settings reported in Section~\ref{sec:experimental_setup}.

\item {\bf Open access to data and code}
    \item[] Question: Does the paper provide open access to the data and code, with sufficient instructions to faithfully reproduce the main experimental results, as described in supplemental material?
    \item[] Answer: \answerYes{}
    \item[] Justification: Code is publicly available at \url{https://github.com/CrazyMint/LEAD}, including the LEAD implementation, top-level training launchers, ablation drivers in \texttt{ablations/}, and MATH preparation scripts in \texttt{scripts/data/}. The README documents installation, training, evaluation settings, and released checkpoints. The MATH dataset is publicly available under its original license. Appendix~\ref{app:protocol_detail} specifies the evaluation protocol, and Appendix~\ref{app:qwen3_lora} describes the additional LoRA study.

\item {\bf Experimental setting/details}
    \item[] Question: Does the paper specify all the training and test details (e.g., data splits, hyperparameters, how they were chosen, type of optimizer) necessary to understand the results?
    \item[] Answer: \answerYes{}
    \item[] Justification: Section~\ref{sec:experimental_setup} and Table~\ref{tab:hyperparams} specify LEAD's optimizer, batch sizes, rollout count, and controller settings. Appendix~\ref{app:protocol_detail} records method-specific configurations and evaluation conventions. Train/test splits and per-benchmark $n$ for avg@$n$ are explicit. Appendix~\ref{app:qwen3_lora} provides the LoRA setup and per-benchmark results.

\item {\bf Experiment statistical significance}
    \item[] Question: Does the paper report error bars suitably and correctly defined or other appropriate information about the statistical significance of the experiments?
    \item[] Answer: \answerYes{}
    \item[] Justification: Table~\ref{tab:seed_summary} reports means and sample standard deviations over three data-order-seed runs. Appendix~\ref{app:uncertainty} provides per-run values, across-run Student-$t$ confidence intervals, and paired question-bootstrap accuracy intervals, with the resampling procedure and sources of variation specified.

\item {\bf Experiments compute resources}
    \item[] Question: For each experiment, does the paper provide sufficient information on the computer resources (type of compute workers, memory, time of execution) needed to reproduce the experiments?
    \item[] Answer: \answerYes{}
    \item[] Justification: Table~\ref{tab:hyperparams} and Appendix~\ref{app:protocol_detail} describe the training configurations. A 1.5B/4K run takes approximately 30 hours on four L40S GPUs. The Qwen3 study uses four L40S GPUs for LEAD and two H200 GPUs for GRPO, taking approximately 114 and 46 hours respectively. Appendix~\ref{app:qwen3_lora} records the matched global training settings and shared evaluation environment.
    
\item {\bf Code of ethics}
    \item[] Question: Does the research conducted in the paper conform, in every respect, with the NeurIPS Code of Ethics \url{https://neurips.cc/public/EthicsGuidelines}?
    \item[] Answer: \answerYes{}
    \item[] Justification: The work uses publicly released benchmarks and openly distributed pretrained model weights under their original licenses, does not involve human subjects, and does not collect or release any new data containing personal information.

\item {\bf Broader impacts}
    \item[] Question: Does the paper discuss both potential positive societal impacts and negative societal impacts of the work performed?
    \item[] Answer: \answerYes{}
    \item[] Justification: Appendix~\ref{app:limitations_impact} discusses lower inference cost, latency, and energy use, along with reduced inspectability of shorter responses and the potential to make harmful uses cheaper. We recommend configurable verbosity in transparency-sensitive applications and retaining the safeguards of the underlying base model.

    
\item {\bf Safeguards}
    \item[] Question: Does the paper describe safeguards that have been put in place for responsible release of data or models that have a high risk for misuse (e.g., pre-trained language models, image generators, or scraped datasets)?
    \item[] Answer: \answerNo{}
    \item[] Justification: We release training code and mathematics-focused fine-tuned checkpoints without additional access restrictions or dedicated model-level safeguards. No new scraped or generated dataset is released. Appendix~\ref{app:limitations_impact} discusses misuse risks and recommends retaining base-model safeguards in deployment.

\item {\bf Licenses for existing assets}
    \item[] Question: Are the creators or original owners of assets (e.g., code, data, models), used in the paper, properly credited and are the license and terms of use explicitly mentioned and properly respected?
    \item[] Answer: \answerYes{}
    \item[] Justification: All third-party assets are cited with their original publications: the verl distributed RL framework (Apache 2.0), the MATH dataset~\cite{hendrycks2021measuring}, the DeepSeek-R1-Distill model family~\cite{guo2025deepseek}, and the math grading utilities from \texttt{deepscaler}. The released code retains the upstream Apache 2.0 LICENSE inherited from verl and is released under the same license.

\item {\bf New assets}
    \item[] Question: Are new assets introduced in the paper well documented and is the documentation provided alongside the assets?
    \item[] Answer: \answerYes{}
    \item[] Justification: The public repository at \url{https://github.com/CrazyMint/LEAD} provides a README covering requirements, installation, training, evaluation settings, results, and licensing. It links released fine-tuned checkpoints, including \url{https://huggingface.co/Kotom1/math_lead_4k_deepseek-r1-1.5b}. The manuscript documents training configurations, evaluation details, and the additional analyses.

\item {\bf Crowdsourcing and research with human subjects}
    \item[] Question: For crowdsourcing experiments and research with human subjects, does the paper include the full text of instructions given to participants and screenshots, if applicable, as well as details about compensation (if any)? 
    \item[] Answer: \answerNA{}
    \item[] Justification: The paper does not involve crowdsourcing or research with human subjects.

\item {\bf Institutional review board (IRB) approvals or equivalent for research with human subjects}
    \item[] Question: Does the paper describe potential risks incurred by study participants, whether such risks were disclosed to the subjects, and whether Institutional Review Board (IRB) approvals (or an equivalent approval/review based on the requirements of your country or institution) were obtained?
    \item[] Answer: \answerNA{}
    \item[] Justification: The paper does not involve research with human subjects.

\item {\bf Declaration of LLM usage}
    \item[] Question: Does the paper describe the usage of LLMs if it is an important, original, or non-standard component of the core methods in this research? Note that if the LLM is used only for writing, editing, or formatting purposes and does \emph{not} impact the core methodology, scientific rigor, or originality of the research, declaration is not required.
    \item[] Answer: \answerNA{}
    \item[] Justification: LLMs are the subject of the experiments (the policy being trained), not a non-standard methodological component: pretrained DeepSeek-R1-Distill and Qwen3 checkpoints are fine-tuned with the proposed RL recipe. No LLM is used as an agent, oracle, judge, or pipeline step in the methodology.

\end{enumerate}

\end{document}